\documentclass[10pt,journal,cspaper,compsoc]{IEEEtran}

\usepackage{booktabs,caption}
\usepackage{graphicx}
\usepackage{mathtools}
\usepackage{enumerate}
\usepackage {dsfont}
\usepackage{booktabs}
\usepackage{times}
\usepackage{epsfig}
\usepackage{balance}
\usepackage{textcomp}
\usepackage{color}
\usepackage{amsmath,amssymb} 
\usepackage{cite}
\usepackage[]{algorithm2e}
\usepackage{multirow}
\usepackage{url}
\usepackage{mathabx}
\usepackage[table]{xcolor}
\usepackage{booktabs,caption,fixltx2e}
\usepackage[flushleft]{threeparttable}
\usepackage{color,soul}
\soulregister\ref7
\soulregister\cite7
\soulregister\citep7
\soulregister\eqref7

\ifCLASSOPTIONcompsoc
\else
\fi
\ifCLASSINFOpdf
\else
\fi
\begin{document}
%
\title{Histogram of Oriented Principal Components for Cross-View Action Recognition}

\author{Hossein~Rahmani,
        Arif~Mahmood,
        Du~Huynh,~\IEEEmembership{Member,~IEEE,}
        and~Ajmal~Mian,~\IEEEmembership{Member,~IEEE}
\IEEEcompsocitemizethanks{\IEEEcompsocthanksitem The authors are with the School of Computer Science and Software Engineering, The University of Western Australia, 35 Stirling Highway, Crawley, Western Australia, 6009.\protect\\
E-mail: hossein@csse.uwa.edu.au, \hfil\break\{arif.mahmood,du.huynh,ajmal.mian\}@uwa.edu.au
}
\thanks{}}

\markboth{Manuscript for Review, 2014}%
{Shell \MakeLowercase{\textit{et al.}}: Bare Demo of IEEEtran.cls for Computer Society Journals}

\IEEEcompsoctitleabstractindextext{%
\begin{abstract}
Existing techniques for 3D action recognition are sensitive to viewpoint variations because they extract features from depth images which are viewpoint dependent. In contrast, we directly process pointclouds for cross-view action recognition from unknown and unseen views. We propose the Histogram of Oriented Principal Components (HOPC) descriptor that is robust to noise, viewpoint, scale and action speed variations. At a 3D point, HOPC is computed by projecting the three scaled eigenvectors of the pointcloud within its local spatio-temporal support volume onto the vertices of a regular dodecahedron. HOPC is also used for the detection of Spatio-Temporal Keypoints (STK) in 3D pointcloud sequences so that view-invariant STK descriptors (or Local HOPC descriptors) at these key locations only are used for action recognition. We also propose a global descriptor computed from the normalized spatio-temporal distribution of STKs in 4-D, which we refer to as STK-D. We have evaluated the performance of our proposed descriptors against nine existing techniques on two cross-view and three single-view human action recognition datasets. The Experimental results show that our techniques provide significant improvement over state-of-the-art methods.


\end{abstract}

}

\maketitle
\IEEEdisplaynotcompsoctitleabstractindextext

\IEEEpeerreviewmaketitle

\section{Introduction}
Human action recognition has numerous applications in smart surveillance, human-computer interaction, sports and elderly care\cite{survey1,survey2}. Kinect like depth cameras have become popular for this task because depth sequences are somewhat immune to variations in illumination, clothing color and texture. However, the presence of occlusions, sensor noise, variations in action execution speed and most importantly sensor viewpoint still make action recognition challenging. 

Designing an efficient representation for 3D video sequences is an important task for many computer vision problems. Most existing techniques ({\it e.g.}\cite{CCD,thau,myLLC,MyWACV14,DMM}) treat depth sequences similar to conventional videos and use color-based action recognition representations. However, simple extensions of color based action recognition techniques to depth sequences are not optimal\cite{DSTIP,HON4D}. Instead of processing depth sequences, richer geometric features can be extracted from 3D pointcloud videos. 


Action recognition research\cite{SNV,thau,4DQuantization,SDPM,DSTIP,HON4D,myLLC,MyWACV14,DMM,MP,eigenjoints} has mainly focused on actions captured from a fixed viewpoint. However, a practical human action recognition system should be able to recognize actions from different views. Some view-invariant approaches\cite{29,26,15,10,4,hankelet,17,21,23,22,3,28,ViewInvariantJoint3D,AL2} have also been proposed for cross-view action recognition where recognition is performed from an unknown and/or unseen view. These approaches generally rely on geometric constraints\cite{29,26,15,10,4}, view-invariant features\cite{hankelet,17,21,23,22,3,28}, and human body joint tracking\cite{ViewInvariantJoint3D,AL2}. More recent approaches transfer features across views\cite{and-or,CTE,SDVI,CVV,8,7,14,virtualviews}. However, these methods do not perform as good as fixed view action recognition. The majority of cross-view action recognition research has focused on color videos or skeleton data. Cross-view action recognition from 3D videos remains an under explored area. We believe that cross-view action recognition from 3D pointcloud videos holds more promise because view-invariant features can be extracted from such videos.

\begin{figure}[t]
\begin{center}
\includegraphics[width=\linewidth]{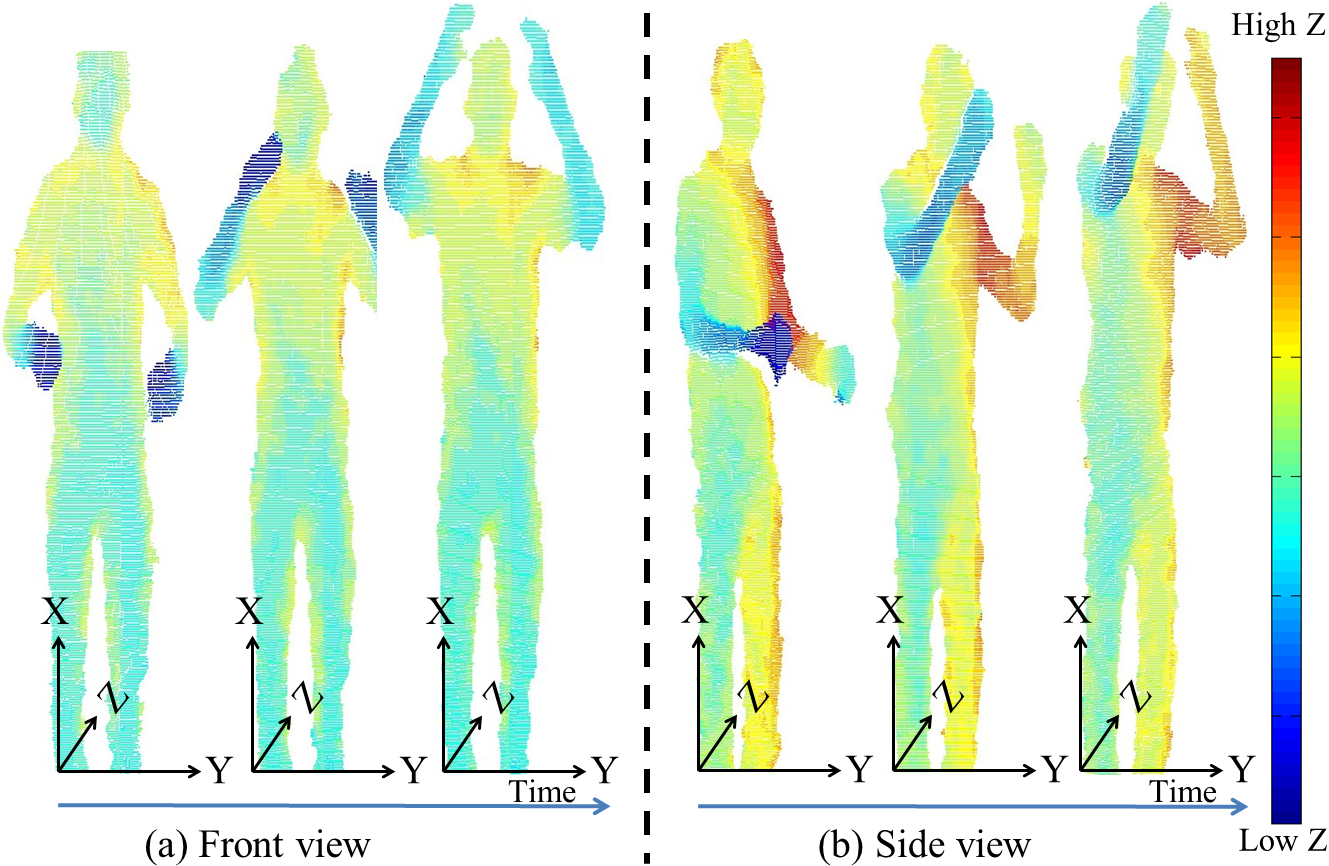}
\end{center}
\caption{\small 3D pointcloud sequences of a subject performing the {\it holding head} action. Notice how the depth values (color) change significantly with viewpoint.}
\label{fig:PCloudSeq}
\end{figure}

We approach the cross-view action recognition problem from a novel perspective by directly processing the 3D pointcloud sequences (Fig.~\ref{fig:PCloudSeq}). We extend our previous research\cite{MyECCV} where we proposed a new descriptor, the Histogram of Oriented Principal Components (HOPC), to capture the local geometric characteristics around each point in a 3D pointcloud sequence. Based on HOPC, we propose a Spatio-Temporal Keypoint (STK) detection method so that view-invariant Local HOPC descriptors are extracted from the most discriminative points within a sequence of 3D pointclouds. We also propose another descriptor, STK-D, which is computed from the spatio-temporal distribution of the STKs. Since Local HOPC and STK-D capture complementary information, their combination significantly improves the cross-view action recognition accuracy over existing state-of-the-art.

\begin{figure}[t]
\begin{center}
\includegraphics[width=\linewidth]{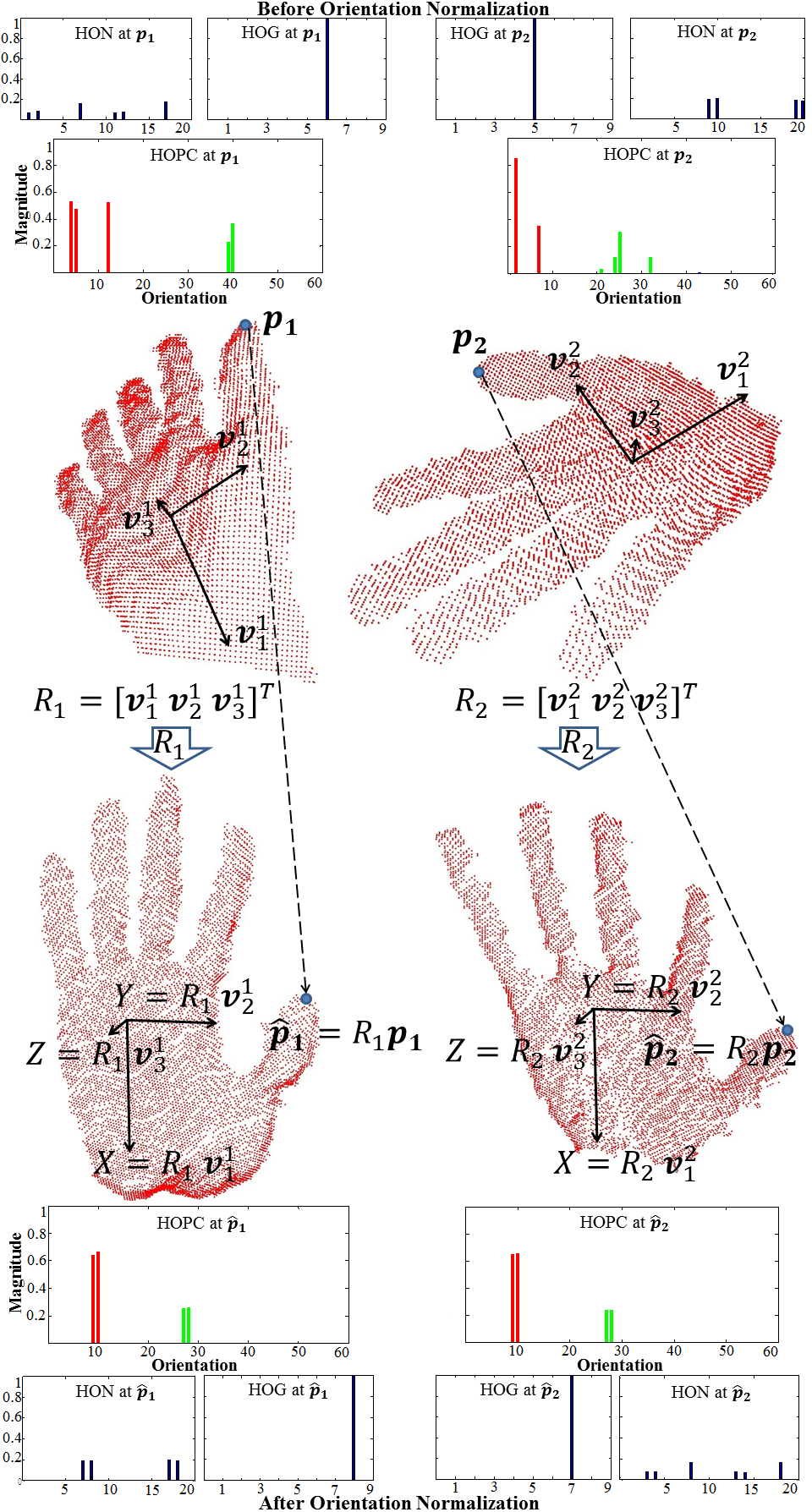}
\end{center}
\caption{\small After orientation normalization, the HOPC descriptors are similar for the two views. However, the HON and HOG descriptors are still different. }
\label{fig:ONorm}
\end{figure}

To achieve view invariance for HOPC, all points within an adaptable spatio-temporal support volume of each STK are aligned along the eigenvectors of its spatial support volume. In other words, the spatio-temporal support volume is aligned in a local object centered coordinate basis. Thus, HOPC descriptor extracted from this aligned support volume is view-invariant (Fig.~\ref{fig:ONorm}). Note that this strategy does not necessarity work for other descriptors as shown in Fig.~\ref{fig:ONorm}. As humans often perform the same action at different speeds, for speed invariance, we propose automatic temporal scale selection that minimizes the eigenratios over a varying temporal window size independently at each STK.

Our four main contributions are summarized as follows: Firstly, we propose the HOPC descriptor which encodes shape and motion in a robust way. Secondly, we propose a view-invariant Spatio-Temporal Keypoint (STK) detector that is integrated with HOPC in the sense that it detects points that are suitable for HOPC. Thirdly, we propose a global action descriptor based on the spatio-temporal distribution of STKs. Finally, we propose a method for viewpoint and speed invariant action recognition. Moreover, we introduce a new UWA3D Multiview Activity II dataset in addition to \cite{MyECCV} which contains $30$ actions performed by $10$ subjects from four different views. This dataset is larger in number of action classes than existing 3D action datasets. 

The proposed descriptors have been evaluated on two multi-view and three single-view human action recognition datasets. The former includes the Northwestern-UCLA Multiview Action3D\cite{and-or} and the UWA3D Multiview Activity II datasets whereas the latter includes MSR Action3D\cite{Bag3DPoints}, MSR Daily Activity3D\cite{ActionLet2012}, and MSR Gesture3D\cite{Wang2012} datasets. Our extensive experimental results show that the proposed descriptors achieved significantly better accuracy compared to the nine existing state-of-the-art techniques\cite{HON4D,SNV,LARP,CCD,NDVV,ViewInvariantJoint3D,virtualviews,AL2,and-or}.

\section{Related Work}
\label{RW}
Based on the data type, action recognition methods can be divided into three categories including color-based, skeleton-based and depth-based methods. In color videos, a significant portion of the existing work has been proposed for single-view action recognition, where the training and test videos are captured from the same view.  In order to recognize actions across different views, one approach is to collect data from all possible views and train a separate classifier for each view. However, this approach does not scale well due to the requirement of a large number of labeled samples for each view and it becomes infeasible as the number of action categories increases. To overcome this problem, some techniques  infer 3D scene structure and use geometric transformations to achieve view invariance \cite{29,26,15,10,4}. These methods critically rely on accurate detection of the body joints and contours, which are still open problems in real-world settings. Other methods focus on spatio-temporal features which are inherently view-invariant\cite{hankelet,17,21,23,22,3,28}. However, these methods have limitations as some of them require access to mocap data while others compromise discriminative power to achieve view invariance\cite{feature}.

More recently, knowledge transfer based methods\cite{and-or,CTE,SDVI,CVV,8,7,14,virtualviews} have become popular. These methods find a view independent latent space in which features extracted from different views are directly comparable. Such methods are either not applicable or perform poorly when the recognition is performed on videos from unknown and, more importantly, from unseen views. To overcome this problem, Wang et al.\cite{and-or} proposed cross-view action representation by exploiting the compositional structure in spatio-temporal patterns and geometrical relations among views. Although their method can be applied to action recognition from unknown and unseen views, it requires 3D skeleton data for training which is not always available. Our proposed approach also falls in this category except that it uses 3D pointcloud sequences and does not require skeleton data. To the best of our knowledge, we are the first to propose cross-view action recognition using 3D pointcloud videos.

In skeleton-based action recognition methods, multi-camera motion capture (MoCap) systems \cite{mocap} have been used for human action recognition. However, such specialized equipment is marker-based and expensive. On the other hand, some other methods\cite {LARP,ActionLet2012,AL2,MP,ViewInvariantJoint3D,eigenjoints} use the human joint positions extracted by the OpenNI tracking framework\cite{SingleDepth}. For example, Yang and Tian\cite{eigenjoints} used pairwise 3D joint position differences in each frame and temporal differences across frames to represent an action. Since 3D joints cannot capture all the discriminative information, the action recognition accuracy is compromised. Wang et al.\ \cite{ActionLet2012} extended this approach by computing the histogram of occupancy patterns of a fixed region around each joint in each frame. In order to make this method more robust to viewpoint variations, they proposed a global orientation normalization using the skeleton data\cite{AL2}. In this method, a plane is fitted to the joints and a rotation matrix is computed to rotate this plane to the $XY$-plane. However, this method is only applicable if the subject is in an upright pose. Moreover, when the subject is in a non-frontal view, the joint positions may have large errors, making the normalization process unreliable. In contrast, our proposed orientation normalization method does not need the joint positions and can efficiently work in non-frontal as well as non upright positions. In addition to that, as our method performs local orientation normalization at each STK, it is more robust than the single global normalization proposed by\cite{AL2}.

%

Many of the existing depth-based action recognition methods use global features such as silhouettes and space-time volume information. For example, Li et al.\ \cite{Bag3DPoints} sampled boundary pixels from 2D silhouettes as a bag of features. Yang et al.\ \cite{DMM} added temporal derivatives of 2D projections to get Depth Motion Maps (DMM). Vieira et al.\ \cite{STOP} computed silhouettes in 3D by using the space-time occupancy patterns. Oreifej and Liu\cite{HON4D} extended histogram of oriented 3D normals \cite{HONV} to 4D by adding the time derivative. Recently, Yang and Tian\cite{SNV} extended HON4D by concatenating the 4D normals in the local neighbourhood of each pixel as its descriptor. Our proposed HOPC descriptor is more informative than HON4D\cite{MyECCV} because it captures the spread of data in three principal directions. Holistic methods may fail in scenarios where the subject significantly changes her/his spatial position\cite{SNV,HON4D}. Some other methods use local features where a set of interest points are extracted from the depth sequence and a local feature descriptor is computed for each interest point. For example, Cheng et al.\cite{CCD} used the Cuboid interest point detector\cite{STIP} and proposed a Comparative Coding Descriptor (CCD). Due to the presence of noise in depth sequences, simply extending color-based interest point detectors, such as Cuboid\cite{Dollar}, 3D Hessian\cite{Hessian} and 3D Harris\cite{STIP}, degrades the efficiency and effectiveness of these detectors as most interest points are detected at irrelevant locations\cite{DSTIP,HON4D}. 

Motion trajectory based action recognition methods\cite{MOTra,DensTraj,Traj,hankelet} are also not reliable in depth sequences\cite{HON4D}. Therefore, recent depth based action recognition methods resorted to alternative ways to extract more reliable interest points. Wang et al.\cite{Wang2012} proposed Haar features to be extracted from each random subvolume. Xia and Aggarwal\cite{DSTIP} proposed a filtering method to extract spatio-temporal interest points. Their approach fails when the action execution speed is faster than the flip of the signal caused by sensor noise. Moreover, both techniques are not robust to viewpoint variations. 


\begin{figure}[t]
\begin{center}
\includegraphics[width=\linewidth]{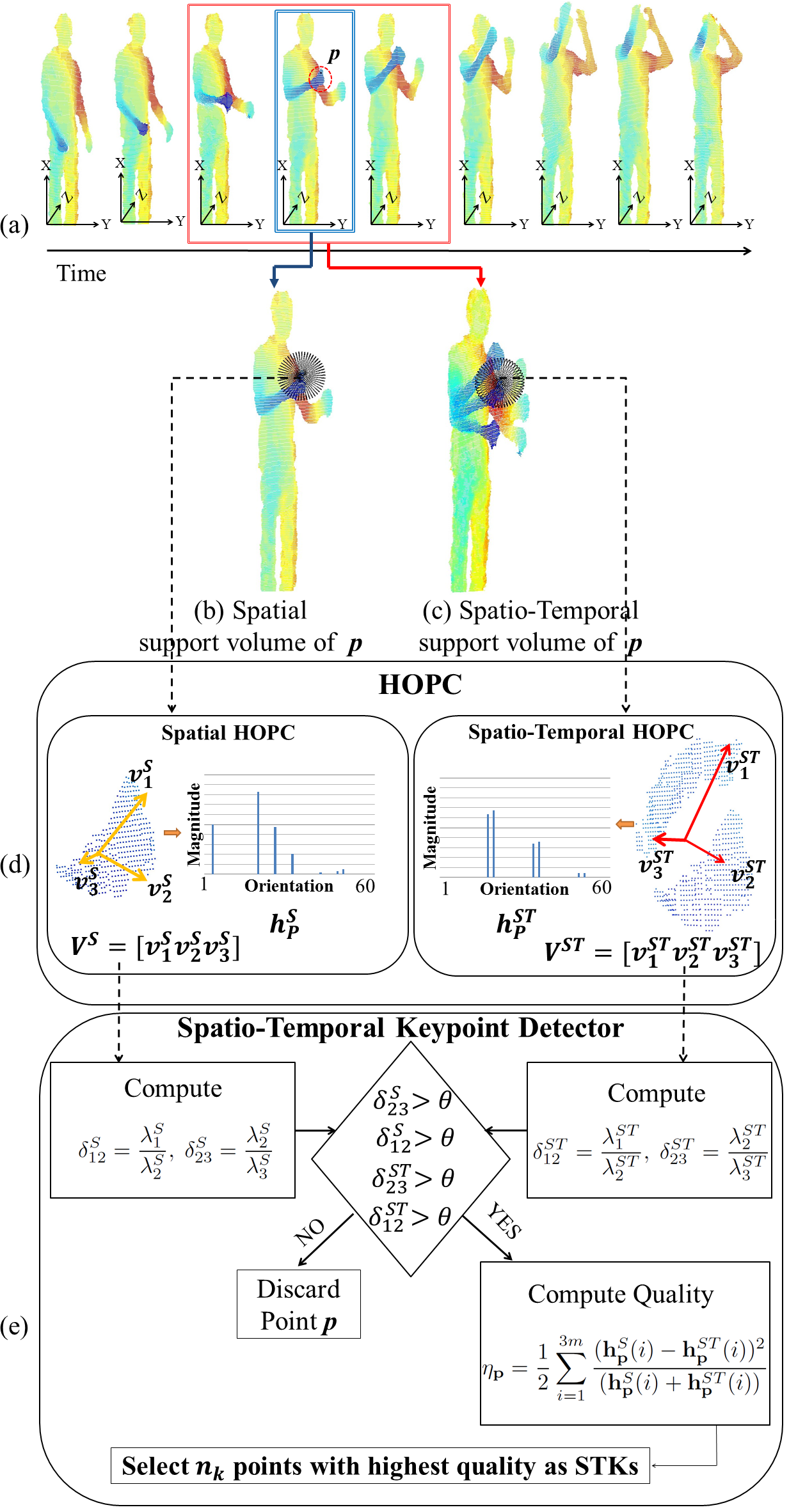}
\end{center}
\caption{\small Spatio Temporal Keypoint (STK) detection. (a) A 3D pointcloud sequence corresponding to the {\it holding head} action, (b) the spatial support volume of a particular point ${\bf p}$, (c) the spatio-temporal support volume of ${\bf p}$, (d) the HOPC descriptors, (e)  STK detection.}
\label{fig:KeyPointAlg}
\end{figure}

\section{HOPC: Histogram of Oriented Principal Components}
\label{ModuleA}
HOPC is extracted at each point within a sequence of 3D pointclouds ${\cal Q}=\text{seq}(Q_1, \cdots,Q_t,\cdots,Q_{n_f})$, where $n_f$ denotes the number of 3D pointclouds in the sequence and $Q_t$ is the 3D pointcloud at time $t$. Consider a point ${\bf p} = (x_t, \; y_t, \; z_t)^{\top}$ in $Q_t$. We define two different support volumes for ${\bf p}$: a {\it spatial support volume} and a {\it spatio-temporal support volume}. The spatial support volume of ${\bf p}$, denoted by $\Omega^{\text{S}}({\bf p})$, contains the 3D points in $Q_t$ that are in a sphere of radius $r$ centered at ${\bf p}$ (Fig. \ref{fig:KeyPointAlg}(b)). To define the spatio-temporal support volume of ${\bf p}$, denoted by $\Omega^{\text{ST}}({\bf p})$, we merge the sequence of pointclouds in the small time interval $[t-\tau, t+\tau]$. The 3D points which are in a sphere of radius $r$ centered at ${\bf p}$ are considered as $\Omega^{\text{ST}}({\bf p})$ (Fig. \ref{fig:KeyPointAlg}(c)). 

The covariance matrix $C^{\alpha}$ of the points ${{\bf q} \in \Omega^{\alpha}({{\bf p}})}, \alpha \in \{\text{ST, S}\}$ is given by:
\begin{equation} C^{\alpha} = \frac{1}{n_p} \sum_{{\bf q} \in \Omega^{\alpha}({{\bf p}})}
{({\bf q} - {\mu}){({\bf q} - {\mu})}^{\top}}, \label{eq:ScatterMatrix} \end{equation}
where \begin{equation*}\; \mu = \frac{1}{n_p} \sum_
{{\bf q} \in \Omega^{\alpha}({{\bf p}})}{\bf q}, \end{equation*}
and $n_p=|\Omega^{\alpha}({{\bf p}})|$ denotes the number of points in the support volume of ${\bf p}$. Performing eigen decomposition on the covariance matrix $C^{\alpha}$ gives us:
\begin{equation} V^{\alpha}E^{\alpha}{V^{\alpha}}^{\top}=C^{\alpha}, \end{equation} 
where $E^{\alpha}$ is a diagonal matrix containing the eigenvalues $\lambda^{\alpha}_1 \geq \lambda^{\alpha}_2 \geq \lambda^{\alpha}_3 \geq 0$ of $C^{\alpha}$ and $V^{\alpha}=[{\bf v}^{\alpha}_1 \; {\bf v}^{\alpha}_2 \; {\bf v}^{\alpha}_3]$ contains the three corresponding orthonormal eigenvectors. 

The HOPC descriptor is built by projecting each eigenvector onto $m$ directions obtained from the vertices of a {\it regular polyhedron}. In particular, we consider a {\it regular dodecahedron} which is composed of $m=20$ vertices, each of which corresponds to a histogram bin. Let $\{ {\bold u}_i \}_{i=1}^{m}$ be the vertices of a {\it regular dodecahedron} and let $U=[{\bold u}_1,{\bold u}_2,\cdots,{\bold u}_{m}] \in \mathds{R}^{3 \!\times\! m}$. For a {\it regular dodecahedron} centered at the origin, these vertices are given as:
\begin{itemize}
\item 8 vertices from  $(\pm1,\pm1,\pm1)$
\item 4 vertices from  $(0,\pm\varphi^{-1},\pm\varphi)$
\item 4 vertices from  $(\pm\varphi^{-1},\pm\varphi,0)$
\item 4 vertices from  $(\pm\varphi,0,\pm\varphi^{-1})$
\end{itemize}
where $\varphi={(1+\sqrt{5})}/{2}$ is the golden ratio. 

Each eigenvector is a direction in the 3D space representing the distribution of point positions in the support volume. Therefore, its orientation has a $180^\circ$ ambiguity. To resolve this orientation ambiguity, we consider the distribution of point vector directions and their magnitudes within the support volume of ${\bf p}$. That is, for each point ${\bf q} \in \Omega^{\alpha}({\bf p})$, we compute ${\bf o}={\bf q}-{\bf p}$ and we determine the sign of each eigenvector ${\bf v}^{\alpha}_j$ as follows:
\begin{equation}
{\bf v}^{\alpha}_j={\bf v}^{\alpha}_j.\text{sign}\left(\sum_{{\bf q} \in \Omega^{\alpha}({\bf p})}{\text{sign}({\bf o}^{\top}{\bf v}^{\alpha}_j)({\bf o}^{\top}{\bf v}^{v}_j})^2 \right),
\label{sign}
\end{equation}
where the `sign' function returns the sign of an input number. Note that the squared projection operation ensures that small projected values, which are often due to noise, are suppressed. If the signs of eigenvectors ${\bf v}^{\alpha}_1, {\bf v}^{\alpha}_2,$ and ${\bf v}^{\alpha}_3$ disagree, {\it i.e.} ${\bf v}^{\alpha}_1 \times {\bf v}^{\alpha}_2 \neq {\bf v}^{\alpha}_3$, we switch the sign of the eigenvector whose $|\sum_{{\bf q} \in \Omega^{\alpha}({\bf p})}{\text{sign}({\bf o}^{\top}{\bf v}^{\alpha}_j)({\bf o}^{\top}{\bf v}^{\alpha}_j})^2|$ value is the smallest. We then project each eigenvector ${\bf v}^{\alpha}_j$ onto $U$ to give us:
\begin{equation}{\bf b}^{\alpha}_j=U^{\top} {\bf v}^{\alpha}_j \in \mathds{R}^{m}, \text{ for } 1 \le j \le 3. \end{equation} 

If ${\bf v}^{\alpha}_j$ perfectly aligns with ${\bf u}_i \in U$, it should vote into only the $i^{\text{th}}$ bin. However, as the ${\bf u}_i$'s are not orthogonal to each other, ${\bf b}^{\alpha}_j$ will have non-zero projection values in other bins as well. To overcome this effect, we quantize the projection values of ${\bf b}^{\alpha}_j$ by imposing a threshold value $\psi$ computed as follows:
\begin{equation}\psi={{\bf u}^{\top}_k} {\bf u}_l=\varphi+\varphi^{-1},\;\;\text{for} \;\;{\bf u}_k,{\bf u}_l \in U,\end{equation} 
where ${\bf u}_k$ and ${\bf u}_l$ are any two {\it neighbouring} vectors in $U$. The quantized vector is then given by
\[ \hat{{\bf b}}^{\alpha}_j(z) = \left\{
  \begin{array}{l l}
    0 & \quad \text{if ${\bf b}^{\alpha}_j(z) \le \psi$}\\
    {\bf b}^{\alpha}_j(z)-\psi & \quad \text{otherwise},
  \end{array} \right.\]
where $1 \leq z \leq m$ denotes a bin number. For the $j^{\text{th}}$ eigenvector, we define ${\bf h}^{\alpha}_j$ to be $\hat{{\bf b}}^{\alpha}_j$ scaled by their corresponding eigenvalue $\lambda^{\alpha}_j$: 

\begin{equation}  {\bf h}^{\alpha}_j=\frac {\lambda^{\alpha}_j \cdot \hat{{\bf b}}^{\alpha}_j}
  {||\hat{{\bf b}}^{\alpha}_j||_2} \in \mathds{R}^{m}, \text{ for } 1 \le j \le 3.
\end{equation} 
We concatenate the histograms of oriented principal components of the three eigenvectors in decreasing order of magnitudes of their associated eigenvalues to form a descriptor for point ${\bf p}$:
\begin{equation} \label{eq:HOPC} {\bf h}^{\alpha}_{\bf p}=\left[{{\bf h}^{\alpha}_1}^{\top}\; {{\bf h}^{\alpha}_2}^{\top}\; {{\bf h}^{\alpha}_3}^{\top}\right] \in \mathds{R}^{3m}. \end{equation}

The spatial HOPC descriptor ${\bf h}^{\text{S}}_{\bf p}$ encodes the shape of the support volume around ${\bf p}$. On the other hand, the spatio-temporal HOPC descriptor ${\bf h}^{\text{ST}}_{\bf p}$ encodes information from both shape and motion. Since the smallest principal component of the local surface is the total least squares estimate of the surface normal\cite{surfaceNormal}, the surface normals encoded in our descriptor are more robust to noise than gradient-based surface normals used in\cite{HONV,HON4D}. Moreover, HOPC additionally encodes the first two eigenvectors which are more dominant compared to the third one. The computation of the spatial and spatio-temporal HOPC descriptors is shown in Fig. \ref{fig:KeyPointAlg}(d). 

\section{Spatio-Temporal Keypoint (STK) Detection}
\label{ModuleB}
The aim of the STK detection is to find points in 3D pointcloud action sequences that satisfy three constraints:

\begin{itemize}
\item {{\it Repeatability}: STKs should be identified with high repeatability in different samples of the same action in the presence of noise and viewpoint changes.} 
\item {{\it Uniqueness}: A unique coordinate basis should be obtained from the neighbourhood of the STKs for the purpose of view-invariant description.}
\item {{\it Significant spatio-temporal variation}: STKs should be detected where the neighbourhood has significant space-time variations.}
\end{itemize}

To achieve these aims, we propose an STK detection technique which has high repeatability, uniqueness and detects points where space-time variation is significant. Consider a point ${\bf p}=(x_t,\; y_t,\; z_t)^{\top}$ within a sequence of 3D pointclouds. We perform eigen decomposition on the spatial and the spatio-temporal covariance matrices $C^{\text{S}}$ and $C^{\text{ST}}$ as described in Section~\ref{ModuleA}. For the first two constraints, we define the following ratios:
\begin{equation}
\delta^\text{S}_{12}=\frac{\lambda^\text{S}_1}{\lambda^\text{S}_2},\; 
\delta^\text{S}_{23}=\frac{\lambda^\text{S}_2}{\lambda^\text{S}_3},\; 
\delta^{\text{ST}}_{12}=\frac{\lambda^{\text{ST}}_1}{\lambda^{\text{ST}}_2},\; 
\delta^{\text{ST}}_{23}=\frac{\lambda^{\text{ST}}_2}{\lambda^{\text{ST}}_3}.\;
\label{eq:ratio4}
\end{equation}

For 3D symmetrical surfaces, the ratio between the first two eigenvalues or last two eigenvalues are very close to $1$. The principal components at such locations are, therefore, ambiguous. Thus, for a point to be qualified as a potential keypoint, the condition 
\begin{equation}
\{\delta^S_{12},\delta^\text{S}_{23},\delta^{\text{ST}}_{12},\delta^{\text{ST}}_{23}\} > \theta_{\text{STK}} = 1+\epsilon_{\text{STK}},\label{eq:eigenratio}\end{equation} 
must be satisfied, where $\epsilon_{\text{STK}}$ is a small margin to cater for noise. This process eliminates ambiguous points and produces a subset of candidate keypoints which can be described uniquely in a local coordinate basis.

Recall that ${\bf h}^\text{S}_{\bf p}$ in~\eqref{eq:HOPC} represents the spatial HOPC and ${\bf h}^{\text{ST}}_{\bf p}$ the spatio-temporal HOPC at point ${\bf p}$. For the third constraint, a {\it quality factor} $\eta_{\bf p}$ is computed for all candidate keypoints:
\begin{equation}
\eta_{\bf p}=\frac{1}{2}\sum_{i=1}^{3m}{\frac{({\bf h}^\text{S}_{\bf p}(i)-{\bf h}^\text{ST}_{\bf p}(i))^2}{({\bf h}^\text{S}_{\bf p}(i)+{\bf h}^\text{ST}_{\bf p}(i))}}~.
\label{eq:eta}\end{equation}
When ${\bf h}^\text{S}_{\bf p}={\bf h}^{\text{ST}}_{\bf p}$, the quality factor is at the minimum value of $\eta_{\bf p}=0$ which basically means that the candidate point ${\bf p}$ has a stationary spatio-temporal support volume. On the other hand, significant variations in space-time change the direction and magnitude of spatio-temporal eigenvectors with respect to the spatial eigenvectors. Thus, $\eta_{\bf p}$ is large when a significant motion occurs in the spatio-temporal support volume.

\begin{figure}
\begin{center}
\includegraphics[width=\linewidth]{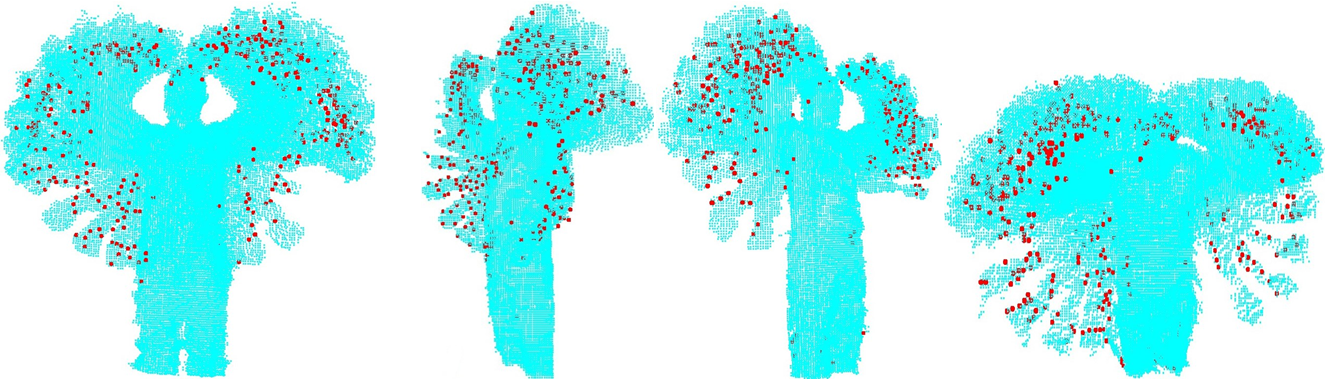}
\end{center}
\caption{\small STKs (shown in red) projected onto $XYZ$ dimensions of all points of a 3D pointcloud sequence corresponding to the {\it two hand waving} action. Four different views are shown. Note that the distribution of STKs encodes the action globally as they are detected only where movement is performed.}
\label{fig:Keypoints}
\end{figure}

STKs that are in the vicinity of each other are similar as they describe more or less the same local support volume. We perform a non-maximum suppression to keep a minimum distance between STKs. We define radius $r'$ (with $r'< r$) and time interval $[t-\tau', t+\tau']$ (with $\tau' \le \tau$) where $t$ is the frame number being considered. The candidate STKs are firstly sorted according to their quality values. Starting from the highest quality STK, all candidate STKs falling within $r'$ and $\tau'$ from it are discarded. The same process is repeated on the remaining candidate STKs until only a desired number, $n_k$, of STKs are left. Figure.~\ref{fig:KeyPointAlg} shows the steps of our STK detection algorithm. Figure.~\ref{fig:Keypoints} shows the extracted STKs from four different views for a 3D pointcloud sequence corresponding to the {\it two hand waving} action. 

\section{View-Invariant STK Description (Local HOPC)}
\label{ModuleC}
The HOPC descriptor discussed in Section~\ref{ModuleA} is not view-invariant yet. We compute Local HOPC only at the STKs since it is possible to normalize the orientation of the local region only at these points {\it i.e.}~a unique local coordinate basis can only be defined at these points. We perform orientation normalization at each STK using the eigenvectors of its spatial covariance matrix $C^\text{S}$ (see Section~\ref{ModuleA}). We consider the eigenvectors $V^\text{S}=[{\bf v}^\text{S}_1\; {\bf v}^\text{S}_2\; {\bf v}^\text{S}_3]$ of $C^{\text{S}}$ as a local object centered coordinate basis. Note that the matrix $V^S$ is orthonormal and can be used as a valid 3D rotation matrix, since: 
\begin{equation}
{\bf v}_i^\text{S}.{\bf v}_j^\text{S} = \left\{
\begin{array}{l l}
1 & \quad \text{if $i=j$}\\
    0 & \quad \text{if $i \neq j$}.
\end{array} \right.\end{equation}

We apply the 3D rotation $R={V^{\text{S}}}^{\top}$ to all the mean-centered points $\{ {\bf q}_i \}_{i=1}^{n_p}$ within the spatio-temporal support volume of ${\bf p}$, $\Omega^{\text{ST}}({\bf p})$, and bring them to a canonical coordinate system:

\begin{equation}{\bf q}'_i=R{\bf q}_i, \;\text{for}\; i=1,\cdots,n_p,\label{eq:orientation}\end{equation}
where ${\bf q}'_i$ denotes the rotated point in the local object centered coordinate basis. Note that the first, second, and third principal components are now aligned with the $X,Y \text{, and}\; Z$ axes of the Cartesian coordinates. Since the same STKs in two different views have the same canonical representation, we can do cross-view keypoint matching (Fig.~\ref{fig:ONorm}). It is important to note that our STK detection algorithm has already pruned ambiguous points to make the local object centered coordinate basis unique, {\it i.e.}~no two eigenvectors have the same eigenvalues. Therefore, the eigenvector with the maximum eigenvalue will always map to the $X$ axis, the second largest to the $Y$ axis and the smallest to the $Z$ axis.

After the orientation normalization given in~\eqref{eq:orientation}, for each point ${\bf q}' \in \Omega^{\text ST}({\bf p})$, we inspect the eigenratios $\delta^{\text{ST}}_{12}$ and $\delta^{\text{ST}}_{23}$ (Eq.~\eqref{eq:eigenratio}) computed using neighbouring points of ${\bf q}'$ to determine how the HOPC descriptor at ${\bf q}'$ should contribute to the STK descriptor computation of ${\bf p}$. When $\delta^{\text{ST}}_{12}$ and $\delta^{\text{ST}}_{23}$ are both larger than $\theta_{\text l} = 1+\epsilon_{\text l}$, where $\epsilon_{\text l}$ is a small margin, all the eigenvectors are uniquely defined and, therefore, can all contribute to the STK descriptor. When $\delta^{\text{ST}}_{12} \le \theta_{\text l}$ and $\delta^{\text{ST}}_{23} > \theta_{\text l}$, the first two eigenvectors are ambiguous and so only the third eigenvector should contribute to the STK descriptor. A similar argument applies to the case where $\delta^{\text{ST}}_{12} > \theta_{\text l}$ and $\delta^{\text{ST}}_{23} \le \theta_{\text l}$. When both $\delta^{\text{ST}}_{12} \le \theta_{\text l}$ and $\delta^{\text{ST}}_{23} \le \theta_{\text l}$, then ${\bf q}'$ has no contribution to the descriptor computation. In summary, the following three criteria need to be considered for the construction of ${{\bf h}^{ST}_{{\bf q}'}}$:\\

\begin{enumerate}
\item {\bf If} $\delta^{\text{ST}}_{12} > \theta_{\text l}$ {\bf \&} $\delta^{\text{ST}}_{23} > \theta_{\text l}$,  ${\bf
  h}^{\text{ST}}_{{\bf q}'}=[{{\bf h}^{\text{ST}}_1}^{\top}\;\;{{\bf h}^{\text{ST}}_2}^{\top}\;\; {{\bf h}^{\text{ST}}_3}^{\top}]^\top$;\vspace{+3mm}
\item {\bf If} $\delta^{\text{ST}}_{12} \le \theta_{\text l}$ {\bf \&} $\delta^{\text{ST}}_{23} > \theta_{\text l}$,
  ${\bf h}^{\text{ST}}_{{\bf q}'}=[{\bf 0}^{\top}\;\;{\bf 0}^{\top}\;\; {{\bf h}^{\text{ST}}_3}^{\top}]^\top$;\vspace{+3mm}
\item {\bf If} $\delta^{\text{ST}}_{12} > \theta_{\text l}$ {\bf \&} $\delta^{\text{ST}}_{23} \le \theta_{\text l}$, 
  ${\bf h}^{\text{ST}}_{{\bf q}'}=[{{\bf h}^{\text{ST}}_1}^{\top}\;\;{\bf 0}^{\top}\;\; {\bf
    0}^{\top}]^\top$.\vspace{+3mm}
\end{enumerate}

Next, the orientation normalized spatio-temporal support volume around the STK ${\bf p}$ is partitioned into $\gamma= n_x \times n_y \times n_t$ spatio-temporal cells along the $X$, $Y$, and $T$ dimensions. We use $c_s, \text{where}\; s=1 \cdots \gamma$, to denote the $s^{\text{th}}$ cell. The cell descriptor ${\bf h}_{c_s}$ is computed by accumulating the ${{\bf h}^{ST}_{{\bf q}'}}$'s
\begin{equation}{\bf h}_{c_s}=\sum_{{\bf q}' \in c_s}{{\bf h}^{ST}_{{\bf q}'}},\label{eq:acc}\end{equation} 
and then normalizing
\begin{equation}{\bf h}_{c_s}\leftarrow \frac{{{\bf h}_{c_s}}}{{||{\bf h}_{c_s}||_2}}.\label{eq:norm}\end{equation}

We define the final view-invariant descriptor, ${\bf h}_{v}$, of STK ${\bf p}$ to be
the concatenation of ${\bf h}_{c_s}$ obtained from all the cells:
\begin{equation}
{\bf h}_{v}={[{\bf h}_{c_1}^{\top}\; {\bf h}_{c_2}^{\top}\; \cdots \; {\bf h}_{c_{\gamma}}^{\top}]}^{\top}.\label{eq:concat}\end{equation}

The above steps are repeated for all the STKs. Thus, the STK descriptors encode view-invariant spatio-temporal patterns that will be used for action description.

%

\section{Action Description}
\label{AD}
\subsection{Bag of STK Descriptors}
We represent each sequence of 3D pointclouds by a set of STK descriptors. Inspired by the successful bag-of-words approach for object recognition, we build a codebook by clustering the STK descriptors (${\bf h}_v$) with the K-means algorithm. Clustering is performed over all action descriptors extracted from all training view samples. Thus, the codebook that we learn is not single action or single view specific. For a fair evaluation, we do not use the target test views in codebook learning or any other training task. We consider each cluster as a codeword that represents a specific spatio-temporal pattern shared by the STKs in that cluster. One codeword is assigned to each STK descriptor based on the minimum Euclidean distance. The histogram of codewords is used as an action descriptor. For classification, we use an SVM classifier with the histogram intersection kernel\cite{HIKSVM}.

\subsection{Mining Discriminative Codebooks}
Not all codewords have the same level of discrimination. Some codewords may encode movements that do not offer good discrimination among different actions, {\it e.g.} the sway of the human body. We use the {\it F-score} to find the most discriminative features in the codebook and discard non-discriminative features. The {\it F-score}\cite{Fscore} measures the discrimination of two sets of real numbers. For more than two sets of real numbers, we use the multiset F-score\cite{FScore2} to measure their discrimination. Given the training histogram of codewords $x_k,\text{for}\; k = 1,\cdots,m$, and $l \geq 2$ action classes, if the number of the samples in the $j$th ($1 \leq j \leq l$) class is $n_j$, then the {\it F-score} of the $i$th histogram bin is defined as:
\begin{equation}
\small
F_i=\frac{\sum_{j=1}^{l}\left({\bar{x}}_i^{(j)}-{\bar{x}}_i \right)^2}{\sum_{j=1}^{l}\frac{1}{n_j-1}\sum_{k=1}^{n_j}{\left({\bar{x}}_{k,i}^{(j)} -{\bar{x}}_i^{(j)}\right)^2}},
\end{equation}
where $\bar{x}_i$ and $\bar{x}_i^{(j)}$ are the average of the $i$th histogram bin of all samples and the $j$th class samples, respectively, and $\bar{x}_{k,i}^{(j)}$ is the $i$th histogram bin of the $k$th sample in the $j$th class. The larger is the {\it F-score}, the more discriminative is the corresponding histogram bin. Therefore, we rank the codewords by their {\it F-scores} and select the codewords whose {\it F-scores} are higher than a  threshold. In our experiments, up to $1.5\%$ improved accuracy was observed by selecting the top $98\%$ discriminative features out of the total $1500$.

\subsection{Encoding Spatio-Temporal STK Distribution}
\label{ModuleD}

The bag-of-words approach efficiently encodes the local spatio-temporal information in a 3D pointcloud sequence. However, it ignores the spatio-temporal relationship among the STKs. We observed that encoding the distribution of STKs in space-time (Fig.~\ref{fig:Keypoints}) can further improve the discrimination between different actions in addition to the bag-of-words based descriptors. To incorporate the space-time positional information of STKs, we propose a method that encodes this information.

Let ${\cal P} = \{ {\bf p}_i \in \mathds{R}^{4}, i=1, \cdots, n_k \}$ represent the set of all selected STKs within a sequence of 3D pointclouds ${\cal Q}$, where $n_k$ is the number of STKs and ${\bf p}_i=(x,\; y,\; z,\; t)^{\top}$ are the coordinates of an STK in the 4D space with $x$ and $y$ being the spatial coordinates, $z$ being depth and $t$ being time. To cope with the heterogeneity in the vectors, we normalize the vectors so that all their components have zero-mean and unit variance.

To simplify the description, let us assume that the set ${\cal P} = \{ {\bf p}_i \in \mathds{R}^{4} \}$ now have all the normalized vectors as described above. By dropping the time axis, we have a set of normalized 3D STKs: ${\cal P}' = \{ {\bf p}'_i \in \mathds{R}^{3} \}$. Eigen decomposition is then applied to the covariance matrix of points in ${\cal P}'$ to yield two eigenratios $\bar{\delta}_{12}=\lambda_1/\lambda_2$ and $\bar{\delta}_{23}=\lambda_2/\lambda_3$. To get a unique coordinate basis, we require that $\bar{\delta}_{12}, \bar{\delta}_{23} >\theta_{\text g} = 1+\epsilon_{\text g}$, where $\epsilon_{\text g}$ is a small constant.
If these constraints are not satisfied, we perform an iterative refinement of STKs as follows. Given $n_k$ initial STKs, in each iteration, $m_k$ (where $m_k \ll n_k$) STKs with the lowest quality factor (Eq.~\eqref{eq:eta}) are removed. Eigen decomposition is applied to the remaining points to yield two new eigenratios. This process is iterated until the eigenratio constraints are satisfied. Generally, three iterations are sufficient.

To achieve a view-invariant representation, all points in ${\cal P}'$ are aligned along $V$, {\it i.e.}, for all ${\bf p}' \in {\cal P}'$, we set ${\bf p}' \leftarrow V^{\top}{\bf p}'$ where $V$ is the eigenvector matrix obtained from the eigen decomposition in the last iteration. The normalized temporal dimension is then reattached to each point in ${\cal P}'$:
\begin{equation}
\widehat{\bf p} \leftarrow [{\bf p}',\;t],
\end{equation}
to form the set $\widehat{\cal P}=\{ {\widehat{\bf p}}_i \in \mathds{R}^{4} \}$. To encode the distribution of STKs in the 4D space, we consider a 4D regular geometric object called {\it polychoron}\cite{4D} which is a 4D extension of the 2D {\it polygon}. The vertices of a {\it regular polychoron} divide the 4D space uniformly, and therefore, each vertex can be considered as a histogram bin. In particular, from the set of {\it regular polychorons}, we consider the {\it 120-cell} {\it regular polychoron} with $600$ vertices as given in Table~\ref{tab:vertices}\cite{4D}.

\begin{table} \normalsize
\caption{\small The 600 vertices of a {\it 120-cell regular polychoron} centered at the origin generated from {\em all} and {\em even} permutations of these coordinates\cite{4D}.}
\setlength{\tabcolsep}{3pt}
   \centering  \begin{tabular}{ccc}
    \toprule
    {Vertices}\;\;   & {Permutation}\;\;  & {Coordinate points} \;\;\\
    \midrule
    \midrule
    $24$\;\;  & {all}\;\; & $0,0,\pm2,\pm2$\;\; \\
    $64$\;\;  & {all}\;\; & $\pm1,\pm1,\pm1,\pm{\sqrt{5}}$\;\; \\
    $64$\;\;  & {all}\;\; & $\pm{\varphi^{-2}},\pm{\varphi},\pm{\varphi},\pm{\varphi}$\;\; \\
    $64$\;\;  & {all}\;\; & $\pm{\varphi^{-1}},\pm{\varphi^{-1}},\pm{\varphi^{-1}},\pm{\varphi^{+2}}$\;\; \\
    $96$\;\;  & {even}\;\; & $0,\pm{\varphi^{-2}},\pm1,\pm{\varphi^{+2}}$\;\; \\
    $96$\;\;  & {even}\;\; & $0,\pm{\varphi^{-1}},\pm{\varphi},\pm{\sqrt{5}}$\;\; \\
    $192$\;\;  & {even}\;\; & $\pm{\varphi^{-1}},\pm1,\pm{\varphi},\pm2$\;\; \\
    \bottomrule
    \end{tabular}
  \label{tab:vertices}
\end{table}

Given the set $\widehat{\cal P}$ constructed above, we project each orientation normalized $\widehat{\bf p}_i$ onto the 600 vertices of the {\it polychoron} and select the vertex with the highest projection value. The histogram bin corresponding to the selected vertex is incremented by one. We repeat this process for all STKs in $\widehat{\cal P}$ and the final histogram is a 600 dimensional STK Distribution \mbox{(STK-D)} descriptor which encodes the global spatio-temporal distribution of STKs of the sequence ${\cal Q}$ in a compact and discriminative form. 

%
%

\section{Adaptable Support Volume}
\label{ASV}
So far, for STK detection and description, we have used a fixed spatio-temporal support volume with spatial radius $r$ and temporal scale $\tau$. However, subjects may have different scales (height and width) and may perform actions at different speeds. Therefore, simply using a fixed spatial radius $r$ and temporal scale $\tau$ is not optimal. Moreover, a  larger value of $r$ enables the proposed descriptor to encapsulate more information about shape but makes the descriptor vulnerable to occlusions. Similarly, a small $\tau$ is preferable over large $\tau$ for better temporal action localization. However, a small $\tau$ may not capture sufficient information about an action if it is performed slowly.

\subsection{Spatial Scale Selection}
Several automatic spatial scale selection methods have been proposed for 3D object retrieval \cite{3DkeypointSurvey}. We adapt the method proposed by Mian et al.\cite{Mian} in object retrieval for action recognition in 3D pointcloud sequences. Note that in the human action recognition problem, the subject's height is available in most cases (which is not the case for object retrieval). Where available, we use the subject's height ($h_s$) to find an appropriate spatial scale. We select the ratios as $r=\sigma h_s,$ where $0<\sigma<1$ is a constant factor. We have empirically selected the value of $\sigma$ to maximize the descriptiveness and robustness of our descriptor to occlusions. In all experiments, we use a fixed value of $\sigma$ for all actions, views and datasets. In our experiments in Section~\ref{Experiments}, we observe that this simple approach achieves almost the same accuracy as the automatic spatial scale selection method adapted from\cite{Mian}. Once we have selected an appropriate spatial scale $r$, then we proceed to select an appropriate temporal scale $\tau$.

\subsection{Automatic Temporal Scale Selection}
\label{ATSS}
\begin{figure}
\begin{center}
\includegraphics[width=\linewidth]{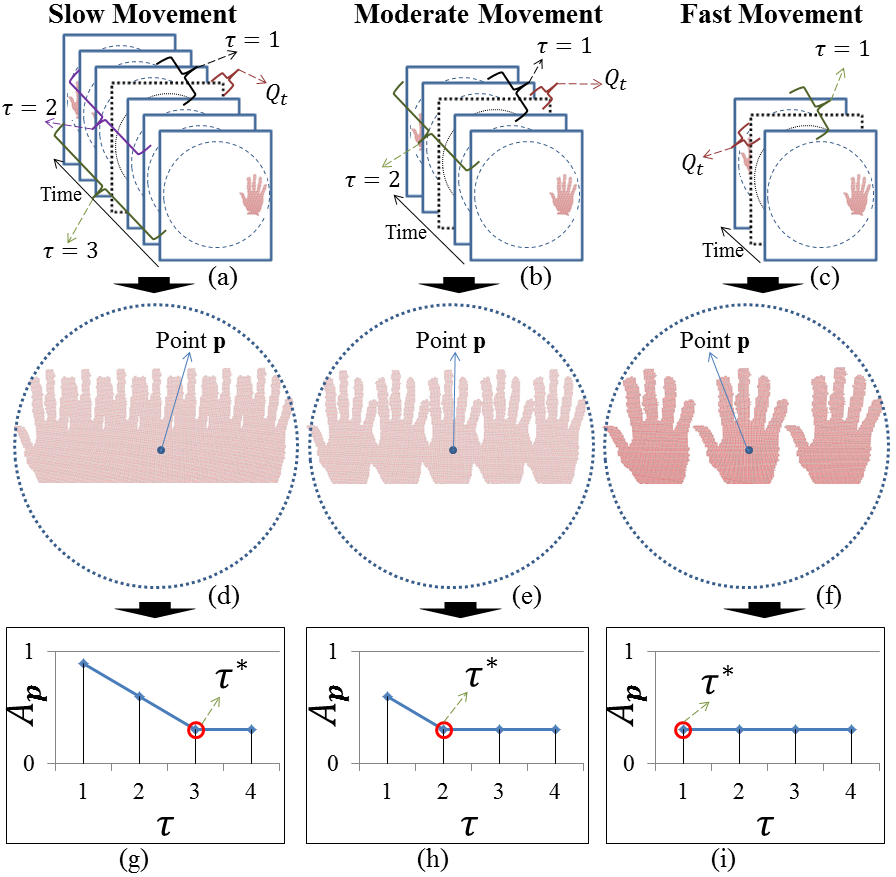}
\end{center}
\caption{\small The same action (hand waving) is shown at three different speeds: (a) slow, (b) moderate, and (c) fast. The number of frames reduces as the action speed increases. For the slow movement, the optimal temporal scale is found to be $\tau^*=3$, for moderate movement $\tau^*=2$, and for fast movement $\tau^*=1$.}
\label{fig:TSS}
\end{figure}

Most existing action recognition techniques\cite{myLLC,HON4D,DSTIP,MyWACV14,STOP,Dollar} use a fixed temporal scale. We observe that variations in action execution speed cause significant disparity among the descriptors from the same action (Fig.~\ref{fig:TSS}). To make our descriptor robust to action speed variations, we propose an automatic temporal scale selection technique. 

Let ${\cal Q}=\text{seq}(Q_1,\cdots,Q_t,\cdots,Q_{n_f})$ be a sequence of 3D pointclouds. For a given point ${\bf p}=(x_t,\;y_t,\;z_t)^{\top} \in Q_t$ and a given temporal scale $\tau$, we can define the spatial support volume $\Omega^{\text{S}}({\bf p})$ and spatio-temporal support volume $\Omega^{\text{ST}}_{\tau}({\bf p})$ of ${\bf p}$.
The covariance matrix of the points falling within $\Omega^{\text{ST}}_{\tau}({\bf p})$ can be eigen-decomposed to yield the eigenvalues $\lambda_1^{\tau} \geq \lambda_2^{\tau} \geq \lambda_3^{\tau} \geq 0$. These steps are similar to those described in Section~\ref{ModuleA}, except that, in this Section, we repeat these steps for each temporal scale $\tau=1,\cdots,\tau_m$, where $\tau_m$ is a fixed upper threshold. For each $\tau$ value, we calculate:

\begin{equation}
A_{{\bf p}}(\tau)=\frac{\lambda_2^{\tau}}{\lambda_1^{\tau}}+\frac{\lambda_3^{\tau}}{\lambda_2^{\tau}}.
\end{equation}
The optimal temporal scale $\tau^*({\bf p})$ for the given point ${\bf p}$ is chosen to be the one that minimizes $A_{\bf p}$ over the range $1 \le \tau \le \tau_m$, {\it i.e.},

\begin{equation}
\tau^*({\bf p})=\underset{\tau}{\operatorname{argmin}}\; A_{\bf p}(\tau).
\end{equation} 

As an example to illustrate this automatic temporal scale selection process, Fig.~\ref{fig:TSS}(a)-(c) show the temporal sequences of pointclouds for the hand waving action performed at three different speeds. The dotted circle shows the sphere defined by the spatial radius $r$ in each pointcloud. The spatial radius $r$ in the three cases is the same because of similar geometry. Our aim here is to select the optimal temporal scale for the point ${\bf p}$ in the pointcloud $Q_t$ shown in black and dotted outline. Figure~\ref{fig:TSS}(d) shows the union of points in the range $Q_{t-3} \cdots Q_{t+3}$ which are within the radius $r$ measured from the coordinate ($x,y,z$) of point ${\bf p}$. Figure~\ref{fig:TSS}(e) and (f) show the union of points in the same way for $Q_{t-2} \cdots Q_{t+2}$ and $Q_{t-1} \cdots Q_{t+1}$, respectively. Figure~\ref{fig:TSS}(g)-(i) show the plots of $A_{\bf p}$ with the variation of $\tau$. Increasing $\tau$ beyond a certain value does not affect the accumulated pointcloud as the value of $A_{\bf p}$ becomes constant. In most cases, increasing $\tau$ decreases $A_{\bf p}$ until a fixed value is reached. We compute $A_{\bf p}(\tau)$ for all values of $\tau$ and find the global minimum  $\tau^*$. When more than one $\tau^*$ exist, the smallest value of $\tau^*$ is chosen.


For each STK, the temporal scale is selected independently and may vary from one STK to the other in the same 3D pointcloud sequence. The proposed temporal scale selection is detailed in Algorithm~\ref{alg:alg3}. The algorithm outputs two variables $\tau^*$ and {\it flag} $\in \{0,1\}$. If the optimal $\tau^*$ is equal to $\tau_m$ then the {\it flag} is set to $0$, indicating that the STK ${\bf p}$ should be discarded. If the computed optimal $\tau^*$ is smaller than $\tau_m$ then the {\it flag} is set to $1$, indicating that the return $\tau^*$ value is the optimal temporal scale for ${\bf p}$.

\RestyleAlgo{algoruled}
\LinesNumbered
\begin{algorithm}[t]

\SetKwData{Left}{left}\SetKwData{This}{this}\SetKwData{Up}{up}
\SetKwFunction{Union}{Union}\SetKwFunction{FindCompress}{FindCompress}
\SetKwInOut{Input}{input}\SetKwInOut{Output}{output}
\Input{$\cal Q$, ${\bf p}$, $r$, and $\tau_{m}$.}
\Output{$\tau^*$, {\it flag}.}  
 \For{$\tau=1:\tau_{m}$}{
  \BlankLine
  Construct $\Omega^{\text{ST}}_{\tau}({\bf p})$;
  \BlankLine
  
  $\mu_{\tau} \leftarrow \frac{1}{n_{p}} \sum_{{\bf q} \in \Omega_{\tau}({{\bf p}})}{\bf q}$; 
  $C_{\tau} \leftarrow \frac{1}{n_{p}} \sum_{{\bf q} \in \Omega_{\tau}({{\bf p}})}
{({\bf q} - {\mu}_{\tau}){({\bf q} - {\mu}_{\tau})}^{\top}}$;
  \BlankLine
  $V_{\tau} \begin{bmatrix}
  \lambda_1^{\tau} & 0 & 0 \\
  0 & \lambda_2^{\tau} & 0 \\
  0 & 0 & \lambda_3^{\tau}
 \end{bmatrix} V_{\tau}^{\top}=C_{\tau}$;
  \BlankLine
  $A_{{\bf p}}(\tau) \leftarrow \frac{\lambda_2^{\tau}}{\lambda_1^{\tau}}+\frac{\lambda_3^{\tau}}{\lambda_2^{\tau}}$;
  \BlankLine
 }
  $\tau^*=\underset{\tau}{\operatorname{argmin}} \;A_{\bf p}(\tau);$
  
 \eIf{$\tau^*==\tau_m$}{{\it flag}$\leftarrow 0;$}{{\it flag}$\leftarrow 1$;}
 \caption{Automatic Temporal Scale Selection}
 \label{alg:alg3}
\end{algorithm}

\section{Experiments}
\label{Experiments}
We evaluate the proposed algorithm on five benchmark datasets, including two multi-view (Northwestern-UCLA Multiview Action3D\cite{and-or}, and UWA3D Multiview Activity II) and three single-view (MSR Action3D\cite{Bag3DPoints}, MSR Daily Activity3D\cite{ActionLet2012}, and MSR Gesture3D\cite{Wang2012}) datasets. Performance is compared to nine existing action recognition methods including Histogram of Oriented 4D Normals (HON4D)\cite{HON4D}, Super Normal Vector (SNV)\cite{SNV}, Lie Algebra Relative Pairs (LARP)\cite{LARP}, Comparative Coding Descriptor (CCD)\cite{CCD}, Virtual Views (VV)\cite{NDVV}, Histogram of 3D Joints (HOJ3D)\cite{ViewInvariantJoint3D}, Discriminative Virtual Views (DVV)\cite{virtualviews}, Actionlet Ensemble (AE)\cite{AL2}, and AND-OR graph (AOG)\cite{and-or}. The baseline results are obtained using publicly available implementations of CCD\cite{CCD}, VV\cite{NDVV}, DVV\cite{virtualviews}, HON4D\cite{HON4D}, SNV\cite{SNV}, and LARP\cite{LARP} from the respective authors' websites. For the remaining three methods AOG\cite{and-or}, HOJ3D\cite{ViewInvariantJoint3D}, and AE\cite{AL2}, we use our implementations because their codes are not publicly available. For CCD\cite{CCD}, VV\cite{NDVV} and DVV\cite{virtualviews}, we use DSTIP\cite{DSTIP}, which is more robust to 3D sensor noise compared to color-based interest point detectors, to extract and describe the spatio-temporal interest points. Our algorithm is robust to many different parameter settings (see Section~\ref{Parameters}). To help the reader reproduce our results, we provide the parameter values that we used in Table~\ref{tab:parameters}. The UWA3D Multiview Activity II dataset and code will be made publicly available.

To evaluate individual components of the proposed algorithm, we report results for the following four settings:\\

\noindent {\bf Holistic HOPC}: A sequence of 3D pointclouds is divided into $\gamma= 6 \times 5 \times 3$ spatio-temporal cells along the $X$, $Y$, and $T$ dimensions. The spatio-temporal HOPC descriptor ${\bf h}^{\text ST}_{\bf p}$ in~\eqref{eq:HOPC} is computed for each point ${\bf p}$ within the sequence. The cell descriptor is computed using~\eqref{eq:acc} and then normalized using~\eqref{eq:norm}. The final descriptor for the given sequence is a concatenation of all the cell descriptors. We use SVM for classification. Similar to HON4D\cite{HON4D} and SNV\cite{SNV}, our Holistic HOPC is suitable for single-view action recognition\cite{MyECCV} and can handle more inter-class similarities of local motions compared to local methods\cite{HON4D}.\\

\noindent {\bf STK-D}: For each sequence of 3D pointclouds, the histogram of spatio-temporal distribution of STKs is used as the sequence descriptor (Section~\ref{ModuleD}).\\

\noindent {\bf Local HOPC}: For each sequence, STKs are detected using the method proposed in Section~\ref{ModuleB}. The proposed orientation normalization is then applied at each STK neighborhood to extract its view-invariant HOPC descriptor (Section~\ref{ModuleC}). The BoW approach is used to describe the sequence.\\

\noindent {\bf Local HOPC+STK-D}: The bag of STK descriptors and the histogram of spatio-temporal distribution of STKs are concatenated to form the sequence descriptor.

\begin{table} 
\centering \caption{\small Parameters and their values: K: number of codewords, $n_k$: number of STKs , $\theta_{\text{STK}}, \theta_{\text l},\theta_{\text g}$: eigenratio thresholds, $n_x\times n_y \times n_t$: spatio-temporal cells (Section~\ref{ModuleC}), $\tau_m$: maximum temporal scale, $m_k$: iterative refinement (Section~\ref{ModuleD}).}
 \setlength{\tabcolsep}{2.2pt}
    \begin{tabular}{|l|c|c|c|c|c|c|c|c|c|c|}
    \hline
	{\bf Parameter} & $K$ & $n_k$ & $\theta_{\text{STK}}$ & $\theta_{\text{l}}$ & $\theta_{\text{g}}$ & $n_x$ & $n_y$ & $n_t$ &$\tau_m$ &  $m_k$\\  
    \hline 
    {\bf Value} & $1500$ & $400$ & $1.3$ & $1.3$ & $1.3$ & $2$ & $2$ & $3$ &$0.2n_f$ &  $0.05n_k$\\ 
    \hline
    \end{tabular}
  \label{tab:parameters}
\end{table}

\subsection{N-UCLA Multiview Action3D Dataset}
\label{UCLA dataset}
\begin{figure}
\begin{center}
\includegraphics[width=\linewidth]{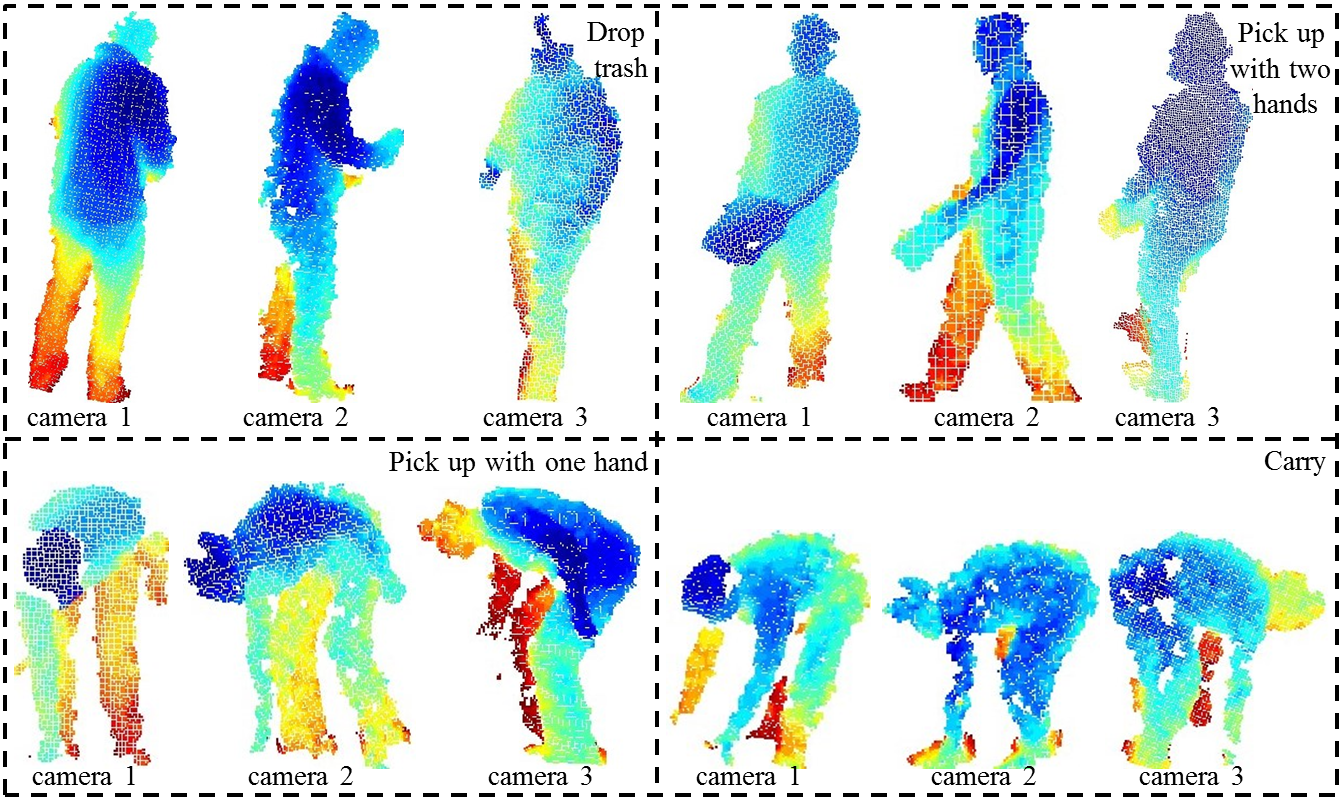}
\end{center}
\caption{\small Sample pointclouds from the Northwestern-UCLA Multiview Action3D dataset\cite{and-or} captured by $3$ cameras.}
\label{fig:UCLAsamples}
\end{figure}

The Northwestern-UCLA dataset\cite{and-or} contains RGB, depth and human skeleton positions captured simultaneously by three Kinect cameras. It consists of 10 action categories: (1) {\it pick up with one hand}, (2) {\it pick up with two hands}, (3) {\it drop trash}, (4) {\it walk around}, (5) {\it sit down}, (6) {\it stand up}, (7) {\it donning}, (8) {\it doffing}, (9) {\it throw}, and (10) {\it carry}. Each action was performed by $10$ subjects $1$ to $6$ times. Figure~\ref{fig:UCLAsamples} shows $12$ sample 3D pointclouds of four actions captured by the three cameras.

\begin{table} 
\centering \caption{\small Comparison of action recognition accuracy (\%) on the Northwestern-UCLA Multiview Action3D dataset where the samples from the first two cameras are used as training data, and the samples from the third camera are used as test data.}
\setlength{\tabcolsep}{8pt}
    \begin{tabular}{lcccccccc}
    \toprule
     Data type & \cellcolor{green!25}RGB  & \cellcolor{red!25}Skeleton & \cellcolor{blue!25}Depth \\
    \midrule \midrule
    CCD\cite{CCD} & \cellcolor{green!25}- & \cellcolor{red!25}- & \cellcolor{blue!25}34.4 \\
    
    VV\cite{NDVV} & \cellcolor{green!25}43.5 & \cellcolor{red!25}- & \cellcolor{blue!25}48.8 \\
    
    HOJ3D\cite{ViewInvariantJoint3D} & \cellcolor{green!25}- & \cellcolor{red!25}54.5 & \cellcolor{blue!25}- \\
    
    DVV\cite{virtualviews} & \cellcolor{green!25}47.8 & \cellcolor{red!25}- & \cellcolor{blue!25}52.1 \\  
      
    AE\cite{AL2} & \cellcolor{green!25}- & \cellcolor{red!25}69.9 & \cellcolor{blue!25}- \\
    
    AOG\cite{and-or} & \cellcolor{green!25}73.3 & \cellcolor{red!25}- &  \cellcolor{blue!25}53.6  \\  
       
    HON4D\cite{HON4D} & \cellcolor{green!25}- & \cellcolor{red!25}- & \cellcolor{blue!25}39.9 \\
    
    SNV\cite{SNV} & \cellcolor{green!25}- & \cellcolor{red!25}- & \cellcolor{blue!25}42.8 \\
    
    LARP\cite{LARP} & \cellcolor{green!25}- & \cellcolor{red!25}74.2 & \cellcolor{blue!25}- \\
     \midrule    
    Holistic HOPC & \cellcolor{green!25}- & \cellcolor{red!25}- & \cellcolor{blue!25}43.4 \\
    
    STK-D & \cellcolor{green!25}- & \cellcolor{red!25}- & \cellcolor{blue!25}53.9 \\
    
    Local HOPC & \cellcolor{green!25}- & \cellcolor{red!25}- & \cellcolor{blue!25}71.9 \\
    
    Local HOPC+STK-D & \cellcolor{green!25}- & \cellcolor{red!25}- & \cellcolor{blue!25}{\bf 80.0} \\
    \bottomrule
    \end{tabular}
  \label{tab:UCLA2}
\end{table}

To compare our method with state-of-the-art algorithms, we use the same experimental setting as\cite{and-or}, using the samples from the first two cameras as training data, and the samples from the third camera as test data. Results are given in Table~\ref{tab:UCLA2}. Holistic approaches such as HON4D\cite{HON4D}, SNV\cite{SNV}, and the proposed Holistic HOPC achieved low recognition accuracy since they are not designed to handle viewpoint changes. Similarly, since depth is a function of viewpoint, CCD\cite{CCD} achieved low accuracy by encoding the differences between the depth values of an interest point and its neighbourhood points.

Among the knowledge transfer based methods, VV\cite{NDVV} and DVV\cite{virtualviews} did not perform well; however, AOG\cite{and-or} obtained high accuracy on only RGB videos. On depth videos, AOG also did not perform well. A possible reason is that depth videos have higher noise levels and interpolating noisy features across views can compromise discrimination ability.

Skeleton based methods such as AE\cite{AL2} and LARP\cite{LARP} achieved high accuracy. We used the scale and orientation normalization of skeletons proposed in LARP\cite{LARP} for AE\cite{AL2} as well which improved the results of AE. However, skeleton data may not be reliable, or even available, when the subject is not in an upright position or is occluded\cite{DSTIP}. More importantly, the application of these methods is limited to human activity recognition where the human skeleton is generally estimated by\cite{SingleDepth}. 

STK-D alone achieved higher accuracy compared to all depth image based methods. This confirms the repeatability of STKs and the robustness of STK-D to viewpoint changes.
The Local HOPC descriptor achieved higher accuracy than STK-D and all depth based methods. Since HOPC and STK-D capture complementary information, their combination (Local \mbox{HOPC+STK-D}) further improved the performance by $8.1$\% achieving the overall best accuracy of $80$\%. Note that this is about $6$\% higher than the nearest competitor LARP\cite{LARP} which requires skeleton data whereas our method does not.

\begin{figure}
\begin{center}
\includegraphics[width=6cm]{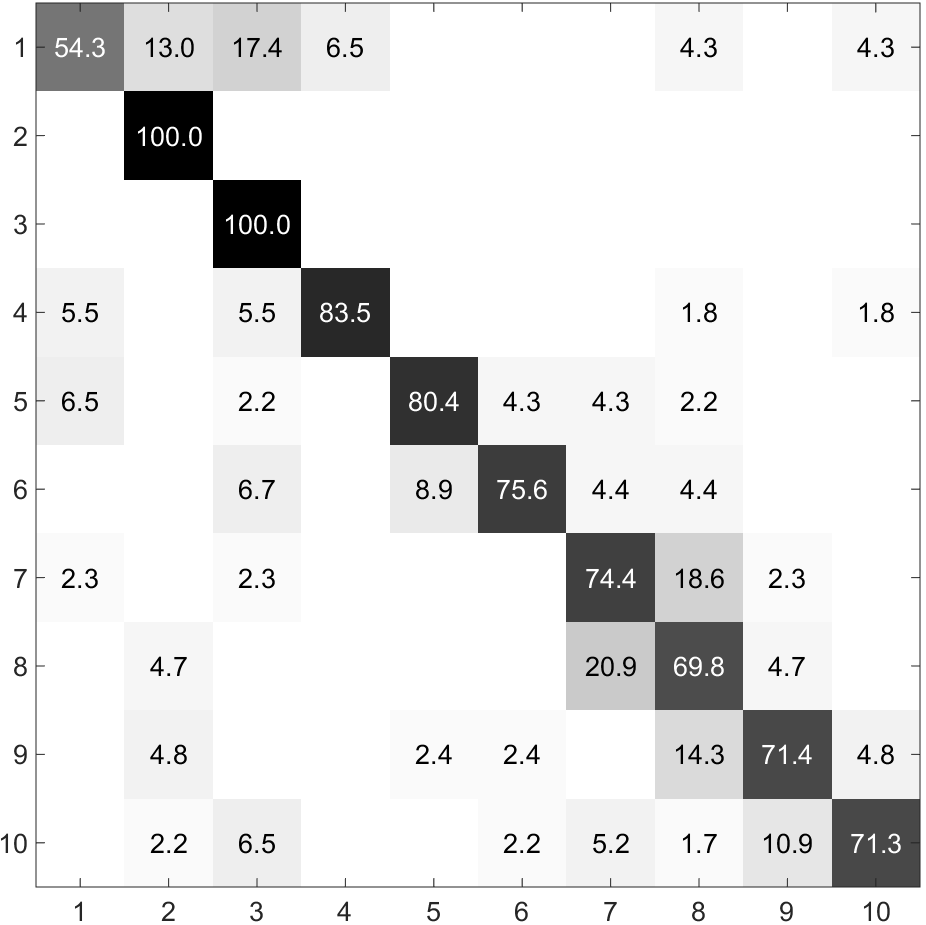}
\end{center}
\caption{\small Confusion matrix of our algorithm on the Northwestern-UCLA Multiview Action3D dataset\cite{and-or}.}
\label{fig:CMUCLA}
\end{figure}

The confusion matrix of our proposed view-invariant Local \mbox{HOPC+STK-D} method is shown in Fig.~\ref{fig:CMUCLA}. The action (7) {\it donning} and action (8) {\it doffing} have maximum confusion with each other because the motion and appearance of these actions are very similar. Similarly, action (1) {\it pick up with one hand} and action (3) {\it drop trash} have high confusion due to similarity in motion and appearance.

\subsection{UWA3D Multiview Activity II Dataset}
\label{MyDataset}
This dataset was collected in our lab using Kinect to emphasize three points: (1) Larger number of human activities. (2) Each subject performed all actions in a continuous manner with no breaks or pauses. Therefore, the start and end positions of body for the same actions are different. (3) Each subject performed the same actions four times while imaged from four different views: front view, left and right side views, and top view. 

\begin{figure}
\begin{center}
\includegraphics[width=\linewidth]{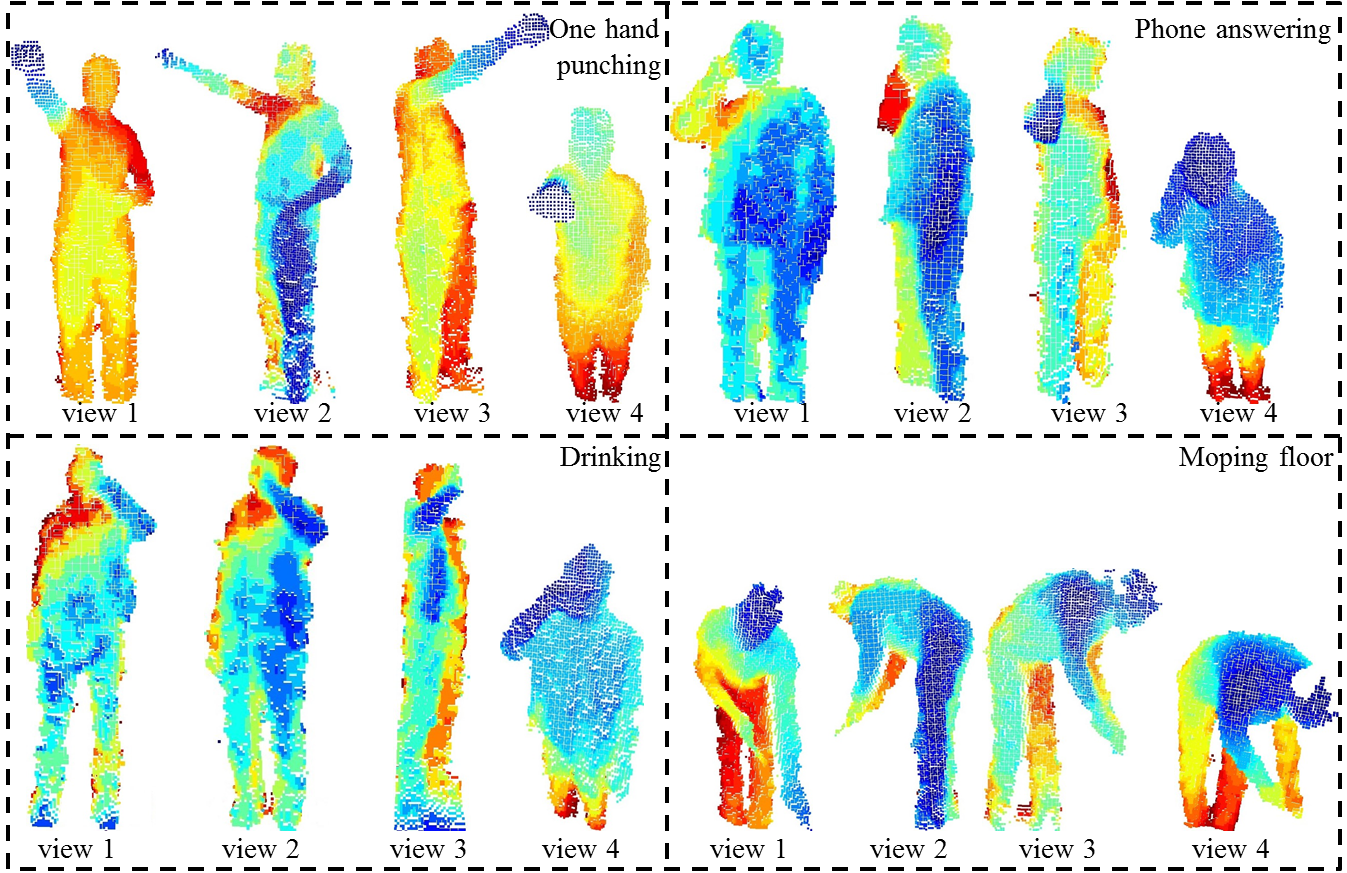}
\end{center}
\caption{\small Sample pointclouds from the UWA3D Multiview Activity II dataset captured by one camera from 4 different views.}
\label{fig:UWA3Dsamples}
\end{figure}

This dataset consists of $30$ human activities performed by $10$ subjects with different scales: (1) {\it one hand waving}, (2) {\it one hand Punching}, (3) {\it two hand waving}, (4) {\it two hand punching}, (5) {\it sitting down}, (6) {\it standing up}, (7) {\it vibrating}, (8) {\it falling down}, (9) {\it holding chest}, (10) {\it holding head}, (11) {\it holding back}, (12) {\it walking}, (13) {\it irregular walking}, (14) {\it lying down}, (15) {\it turning around}, (16) {\it drinking}, (17) {\it phone answering}, (18) {\it bending}, (19) {\it jumping jack}, (20) {\it running}, (21) {\it picking up}, (22) {\it putting down}, (23) {\it kicking}, (24) {\it jumping}, (25) {\it dancing}, (26) {\it moping floor}, (27) {\it sneezing}, (28) {\it sitting down (chair)}, (29) {\it squatting}, and (30) {\it coughing}. To capture depth videos, each subject performed $30$ activities $4$ times in a continuous manner. Each time, the Kinect was moved to a different angle to capture the actions from four different views. Note that this approach generates more challenging data than when actions are captured simultaneously from different viewpoints. We organized our dataset by segmenting the continuous sequences of activities. The dataset is challenging because of varying viewpoints, self-occlusion and high similarity among activities. For example, the actions (16) {\it drinking} and (17) {\it phone answering} have very similar motion, but the location of hand in these two actions is slightly different. Also, some actions such as (10) {\it holding head} and (11) {\it holding back,} have self-occlusion. Moreover, in the top view, the lower part of the body was not properly captured because of occlusion. Figure~\ref{fig:UWA3Dsamples} shows $16$ sample pointclouds of five actions from $4$ views.  

\begin{table*}[t] 
\centering \caption{\small Comparison of action recognition accuracy (\%) on the UWA3D Multiview Activity II dataset. Each time two views are used for training and the remain two views are individually used for testing.}
\setlength{\tabcolsep}{6.5pt}
    \begin{tabular}{lcccccccccccccccccc}
    \toprule
    Training views & \multicolumn{2}{c}{$V_1 \;\&\; V_2$}  & \multicolumn{2}{c}{$V_1 \;\&\; V_3$} & \multicolumn{2}{c}{$V_1 \;\&\; V_4$} & \multicolumn{2}{c}{$V_2 \;\&\; V_3$} & \multicolumn{2}{c}{$V_2 \;\&\; V_4$} & \multicolumn{2}{c}{$V_3 \;\&\; V_4$} &   \multirow{2}{*}{Mean}\\
    \cline{1-13}
    Test view & $V_3$ & $V_4$& $V_2$ & $V_4$ & $V_2$ & $V_3$ & $V_1$ & $V_4$ & $V_1$ & $V_3$ & $V_1$ & $V_2$\\
    \midrule \midrule

     \cellcolor{green!25}AOG\cite{and-or} (RGB) & \cellcolor{green!25}47.3 & \cellcolor{green!25}39.7 & \cellcolor{green!25}43.0 & \cellcolor{green!25}30.5 & \cellcolor{green!25}35.0 & \cellcolor{green!25}42.2 & \cellcolor{green!25}50.7 & \cellcolor{green!25}28.6 & \cellcolor{green!25}51.0 & \cellcolor{green!25}43.2 & \cellcolor{green!25}51.6 & \cellcolor{green!25}44.2 & \cellcolor{green!25}42.3\\ 
     
    \cellcolor{red!25}HOJ3D\cite{ViewInvariantJoint3D} (Skeleton) & \cellcolor{red!25}15.3 & \cellcolor{red!25}28.2 & \cellcolor{red!25}17.3 & \cellcolor{red!25}27.0 & \cellcolor{red!25}14.6 & \cellcolor{red!25}13.4 & \cellcolor{red!25}15.0 & \cellcolor{red!25}12.9 & \cellcolor{red!25}22.1 & \cellcolor{red!25}13.5 & \cellcolor{red!25}20.3 & \cellcolor{red!25}12.7 & \cellcolor{red!25}17.7 \\
    
    
    \cellcolor{red!25}AE\cite{AL2} (Norm. Skeleton) & \cellcolor{red!25}45.0 & \cellcolor{red!25}40.4 & \cellcolor{red!25}35.1 & \cellcolor{red!25}36.9 & \cellcolor{red!25}34.7 & \cellcolor{red!25}36.0 & \cellcolor{red!25}49.5 & \cellcolor{red!25}29.3 & \cellcolor{red!25}57.1 & \cellcolor{red!25}35.4 & \cellcolor{red!25}49.0 & \cellcolor{red!25}29.3 & \cellcolor{red!25}39.8 \\
    
    \cellcolor{red!25}LARP\cite{LARP} (Norm. Skeleton) & \cellcolor{red!25}49.4 & \cellcolor{red!25}42.8 & \cellcolor{red!25}34.6 & \cellcolor{red!25}39.7 & \cellcolor{red!25}38.1 & \cellcolor{red!25}{\bf 44.8} & \cellcolor{red!25}53.3 & \cellcolor{red!25}33.5 & \cellcolor{red!25}53.6 & \cellcolor{red!25}41.2 & \cellcolor{red!25}56.7 & \cellcolor{red!25}32.6 & \cellcolor{red!25}43.4 \\ 
    
    \cellcolor{blue!25}CCD\cite{CCD} (Depth) & \cellcolor{blue!25}10.5 & \cellcolor{blue!25}13.6 & \cellcolor{blue!25}10.3 & \cellcolor{blue!25}12.8 & \cellcolor{blue!25}11.1 & \cellcolor{blue!25}8.3 & \cellcolor{blue!25}10.0 & \cellcolor{blue!25}7.7 & \cellcolor{blue!25}13.1 & \cellcolor{blue!25}13.0 & \cellcolor{blue!25}12.9 & \cellcolor{blue!25}10.8 & \cellcolor{blue!25}11.2 \\
    
    \cellcolor{blue!25}VV\cite{NDVV} (Depth) & \cellcolor{blue!25}20.2 & \cellcolor{blue!25}22.0 & \cellcolor{blue!25}19.9 & \cellcolor{blue!25}22.3 & \cellcolor{blue!25}19.3 & \cellcolor{blue!25}20.5 & \cellcolor{blue!25}20.8 & \cellcolor{blue!25}19.3 & \cellcolor{blue!25}21.6 & \cellcolor{blue!25}21.2 & \cellcolor{blue!25}23.1 & \cellcolor{blue!25}19.9 & \cellcolor{blue!25}20.9 \\
    
    \cellcolor{blue!25}DVV\cite{virtualviews} (Depth) & \cellcolor{blue!25}23.5 & \cellcolor{blue!25}25.9 & \cellcolor{blue!25}23.6 & \cellcolor{blue!25}26.9 & \cellcolor{blue!25}22.3 & \cellcolor{blue!25}20.2 & \cellcolor{blue!25}22.1 & \cellcolor{blue!25}24.5 & \cellcolor{blue!25}24.9 & \cellcolor{blue!25}23.1 & \cellcolor{blue!25}28.3 & \cellcolor{blue!25}23.8 & \cellcolor{blue!25}24.1 \\    
    
    \cellcolor{blue!25}AOG\cite{and-or} (Depth) & \cellcolor{blue!25}29.3 & \cellcolor{blue!25}31.1 & \cellcolor{blue!25}25.3 & \cellcolor{blue!25}29.9 & \cellcolor{blue!25}22.7  & \cellcolor{blue!25}21.9 & \cellcolor{blue!25}25.0 & \cellcolor{blue!25}20.2 & \cellcolor{blue!25}30.5 & \cellcolor{blue!25}27.9 & \cellcolor{blue!25}30.0 & \cellcolor{blue!25}26.8 & \cellcolor{blue!25}26.7 \\ 
    
    \cellcolor{blue!25}HON4D\cite{HON4D} (Depth) & \cellcolor{blue!25}31.1 & \cellcolor{blue!25}23.0 & \cellcolor{blue!25}21.9 & \cellcolor{blue!25}10.0 & \cellcolor{blue!25}36.6 & \cellcolor{blue!25}32.6 & \cellcolor{blue!25}47.0 & \cellcolor{blue!25}22.7 & \cellcolor{blue!25}36.6 & \cellcolor{blue!25}16.5 & \cellcolor{blue!25}41.4 & \cellcolor{blue!25}26.8 & \cellcolor{blue!25}28.9\\
    
    \cellcolor{blue!25}SNV\cite{SNV} (Depth) & \cellcolor{blue!25}31.9 & \cellcolor{blue!25}25.7 & \cellcolor{blue!25}23.0 & \cellcolor{blue!25}13.1 & \cellcolor{blue!25}38.4 & \cellcolor{blue!25}34.0 & \cellcolor{blue!25}43.3 & \cellcolor{blue!25}24.2 & \cellcolor{blue!25}36.9 & \cellcolor{blue!25}20.3 & \cellcolor{blue!25}38.6 & \cellcolor{blue!25}29.0 & \cellcolor{blue!25}29.9\\
       
    \midrule    
    \cellcolor{blue!25}Holistic HOPC (Depth) & \cellcolor{blue!25}32.3 & \cellcolor{blue!25}25.2 & \cellcolor{blue!25}27.4 & \cellcolor{blue!25}17.0 & \cellcolor{blue!25}38.6 & \cellcolor{blue!25}38.8 & \cellcolor{blue!25}42.9 & \cellcolor{blue!25}25.9 & \cellcolor{blue!25}36.1 & \cellcolor{blue!25}27.0 & \cellcolor{blue!25}42.2 & \cellcolor{blue!25}28.5 & \cellcolor{blue!25}31.8\\
    
    \cellcolor{blue!25}STK-D (Depth) & \cellcolor{blue!25}32.8 & \cellcolor{blue!25}25.1 & \cellcolor{blue!25}38.7 & \cellcolor{blue!25}22.7 & \cellcolor{blue!25}23.7 & \cellcolor{blue!25}23.4 & \cellcolor{blue!25}29.0 & \cellcolor{blue!25}19.2 & \cellcolor{blue!25}27.9 & \cellcolor{blue!25}28.0 & \cellcolor{blue!25}24.5 & \cellcolor{blue!25}30.1 & \cellcolor{blue!25}27.1 \\
    
    \cellcolor{blue!25}Local HOPC (Depth) & \cellcolor{blue!25}42.3 & \cellcolor{blue!25}46.5 & \cellcolor{blue!25}39.1 & \cellcolor{blue!25}49.8 & \cellcolor{blue!25}35.0 & \cellcolor{blue!25}39.3 & \cellcolor{blue!25}51.9 & \cellcolor{blue!25}34.4 & \cellcolor{blue!25}57.9 & \cellcolor{blue!25}35.3 & \cellcolor{blue!25}60.5 & \cellcolor{blue!25}36.5 & \cellcolor{blue!25}44.0 \\  
    
    \cellcolor{blue!25}Local HOPC+STK-D (Depth) & \cellcolor{blue!25}{\bf 52.7} & \cellcolor{blue!25}{\bf 51.8} & \cellcolor{blue!25}{\bf 59.0} & \cellcolor{blue!25}{\bf 57.5} & \cellcolor{blue!25}{\bf 42.8} & \cellcolor{blue!25}44.2 & \cellcolor{blue!25}{\bf 58.1} & \cellcolor{blue!25}{\bf 38.4} & \cellcolor{blue!25}{\bf 63.2} & \cellcolor{blue!25}{\bf 43.8} & \cellcolor{blue!25}{\bf 66.3} & \cellcolor{blue!25}{\bf 48.0} & \cellcolor{blue!25}{\bf 52.2} \\ 
    \bottomrule
    \end{tabular}
  \label{tab:MyDataset6}
\end{table*}

For cross-view action recognition, we use the samples from two views as training data, and the samples from the two remaining views as test data. Table~\ref{tab:MyDataset6} summarizes our results. Since this dataset is more challenging compared to the N-UCLA dataset, the performance of all methods drops significantly. Our Holistic HOPC descriptor achieved higher average recognition accuracy than the depth based methods but lower than the methods which use normalized skeleton data. Among the depth based methods, HON4D\cite{HON4D} and SNV\cite{SNV} are the nearest competitors to the Holistic HOPC. The Local HOPC achieved higher accuracy than STK-D and the Holistic HOPC. Combining STK-D with Local HOPC again improved performance by $8.2$\% achieving the overall best performance of $52.2$\%. Note that this is about $9$\% higher than the nearest competitor LARP\cite{LARP} which uses skeleton data. Local \mbox{HOPC+STK-D} achieved the highest accuracy in all combinations of training and test views except one. The accuracy of skeleton based methods is significantly lower on this dataset because the skeleton data is not accurate for some actions such as {\it drinking}, {\it phone answering}, {\it sneezing} or is not available for some actions such as {\it falling down} and {\it lying down}.

Moreover, the overall accuracy of the knowledge transfer based methods VV\cite{NDVV}, DVV\cite{virtualviews}, and AOG\cite{and-or} when depth videos are used as input data is low because motion and appearance of many actions are very similar and the depth sequences have a high level of noise. Therefore, the view dependent local features used in VV\cite{NDVV}, DVV\cite{virtualviews} and the appearance and motion interpolation based method used in AOG\cite{and-or} are not enough to discriminate between actions in the presence of noise.

\begin{figure}
\begin{center}
\includegraphics[width=\linewidth]{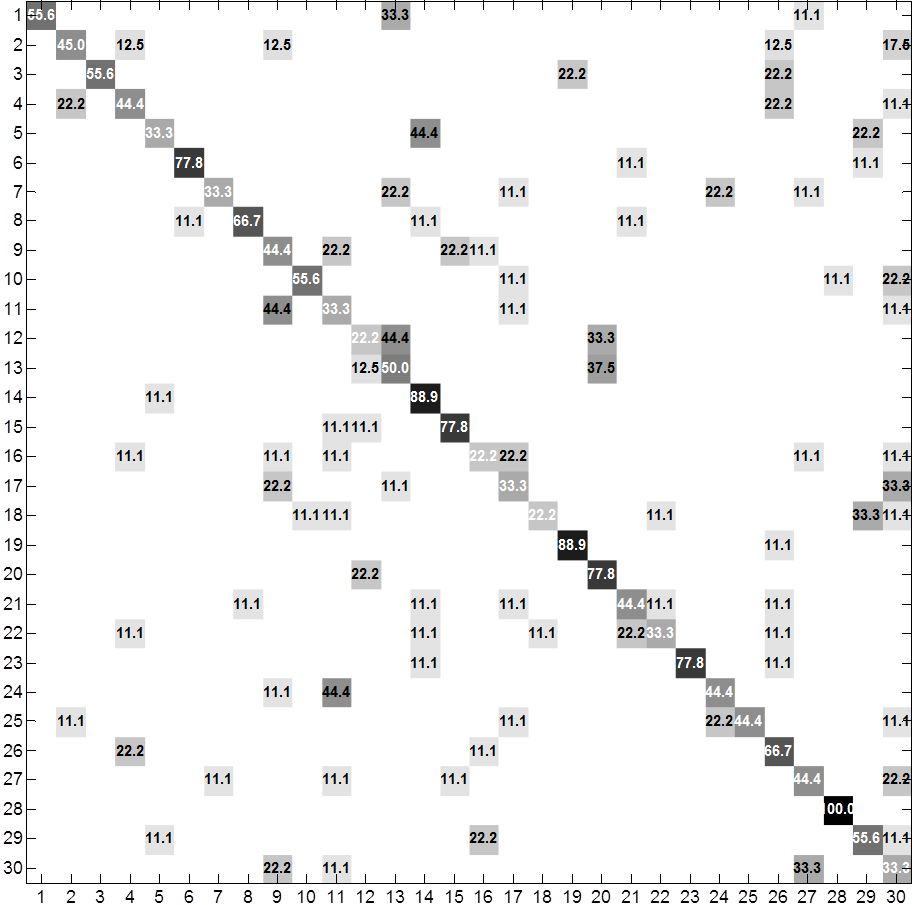}
\end{center}
\caption{\small Confusion matrix of our algorithm on the UWA3D Multiview Activity II dataset when view $V_1$ and view $V_2$ are used for training and view $V_3$ is used for test.}
\label{fig:CMUWA3D}
\end{figure}

Figure~\ref{fig:CMUWA3D} shows the confusion matrix of our proposed view-invariant Local \mbox{HOPC+STK-D} method when videos from view $V_1$ and view $V_2$ are used for training and videos from view $V_3$ are used as test data. The actions that causes the most confusion are (9) {\it holding chest} versus (11) {\it holding back} and (12) {\it walking} versus (13) {\it irregular walking}, because the motion and appearance of these two actions are very similar. 

\subsection{MSR Action3D Dataset}
\label{MSR Action3D dataset}
The MSR Action3D dataset\cite{Bag3DPoints} consists of $20$ actions performed $2$ to $3$ times by $10$ subjects. This dataset is challenging due to high inter-action similarities. Following the protocol of\cite{HON4D}, we use $5$ subjects for training and the remaining $5$ for testing and exhaustively repeated the experiments $252$ folds. Table~\ref{tab:SingleView} compares our algorithms with existing methods. The proposed Holistic HOPC outperformed all methods and achieved $86.5\%$ average accuracy which is more than $2$\% higher than its nearest competitor SNV\cite{SNV} and significantly higher than the skeleton based methods such as HOJ3D\cite{ViewInvariantJoint3D} and LARP\cite{LARP}. The average accuracy of our view-invariant Local \mbox{HOPC+STK-D} method is $82.9\%$ which is still higher than HOJ3D\cite{ViewInvariantJoint3D}, AE~\cite{AL2}, HON4D\cite{HON4D}, and LARP\cite{LARP}.

\subsection{MSR Daily Activity3D Dataset}
\label{MSR Daily Activity3D dataset}
This dataset\cite{ActionLet2012} contains $16$ daily activities performed twice by $10$ subjects, once in standing position and once while sitting. Most activities involve human-object interactions which makes this dataset challenging. We follow the experimental setting of\cite{ActionLet2012} and use samples from half of the subjects as training data, and the rest as test data. 
As shown in Table~\ref{tab:SingleView}, the proposed Holistic HOPC outperformed all techniques achieving an average accuracy of $88.8\%$. The view-invariant Local \mbox{HOPC+STK-D} outperformed AOG\cite{and-or}, HOJ3D\cite{ViewInvariantJoint3D} and LARP\cite{LARP}; however, it achieved lower accuracy than HON4D\cite{HON4D} and SNV\cite{SNV}, because these methods assume that the training and test samples are obtained from the same viewpoint.

\subsection{MSR Gesture3D Dataset}
\label{MSR Gesture3D dataset}
The MSR Gesture3D dataset\cite{Wang2012} contains $12$ American sign language gestures performed $2$ to $3$ times by $10$ subjects. For comparison with previous techniques, we use the leave-one-subject-out cross validation scheme proposed by\cite{Wang2012}. Table~\ref{tab:SingleView} compares our methods to existing ones excluding AE\cite{AL2}, LARP\cite{LARP}, AOG\cite{and-or} and HOJ3D\cite{ViewInvariantJoint3D} since they require 3D joint positions which are not present in this dataset. Our Holistic HOPC outperformed all techniques and achieved an average accuracy of $96.2\%$. The Local \mbox{HOPC+STK-D} achieves an accuracy of $93.6\%$ which is higher than HON4D\cite{HON4D}. 

\begin{table} 
 \caption{\small Comparison of average action recognition accuracy (\%) on the MSR Action3D\cite{Bag3DPoints}, MSR Daily Activity3D\cite{ActionLet2012}, and MSR Gesture3D\cite{Wang2012} datasets. NA: RGB or skeleton data Not Available.}
 \centering
\setlength{\tabcolsep}{3pt}
    \begin{tabular}{lcccccc}
    \toprule
    {Method} & Action & DailyActivity & Gesture\\
    \midrule\midrule
    \cellcolor{green!25}AOG\cite{and-or} (RGB) & \cellcolor{green!25}NA & \cellcolor{green!25}73.1 & \cellcolor{green!25}NA\\
    
    \cellcolor{red!25}HOJ3D\cite{ViewInvariantJoint3D} (Skeleton) & \cellcolor{red!25}63.6 & \cellcolor{red!25}66.8 & \cellcolor{red!25}NA\\  
    
    \cellcolor{red!25}AE\cite{AL2} (Norm. Skeleton) & \cellcolor{red!25}81.6 & \cellcolor{red!25}85.8 & \cellcolor{red!25}NA\\
    
    \cellcolor{red!25}LARP\cite{LARP} (Norm. Skeleton) & \cellcolor{red!25}78.8 &  \cellcolor{red!25}69.4 & \cellcolor{red!25}NA\\
    
    \cellcolor{blue!25}AOG\cite{and-or} (Depth) & \cellcolor{blue!25}NA & \cellcolor{blue!25}53.8 & \cellcolor{blue!25}NA\\ 
     
    \cellcolor{blue!25}HON4D\cite{HON4D} (Depth) & \cellcolor{blue!25}82.2 & \cellcolor{blue!25}80.0 & \cellcolor{blue!25}92.5\\
    
    \cellcolor{blue!25}SNV\cite{SNV} (Depth) & \cellcolor{blue!25}84.4 & \cellcolor{blue!25}86.3 & \cellcolor{blue!25}94.7\\
    
    \midrule
    \cellcolor{blue!25}Holistic HOPC (Depth) & \cellcolor{blue!25}{\bf 86.5} & \cellcolor{blue!25}{\bf 88.8} & \cellcolor{blue!25}{\bf 96.2}\\
    
    \cellcolor{blue!25}Local HOPC+STK-D (Depth) & \cellcolor{blue!25}82.9 & \cellcolor{blue!25}78.8 & \cellcolor{blue!25}93.6\\
    \bottomrule
    \end{tabular}
  \label{tab:SingleView}
\end{table}

\subsection{Effects of Adaptable Support Volume}
\subsubsection{Spatial Scale Selection}
In this experiment, we evaluate the influence of three different approaches for spatial scale selection at each STK. In the first approach, we use a constant spatial scale for all subjects. In the second approach, we select a scale for each subject relative to the subject's height. In the third one, we use the automatic spatial scale selection method proposed by Mian et al.\cite{Mian}. Table~\ref{tab:spatialscale} shows the average accuracy of the proposed method in the three settings. Using the subject's height to find a subject specific scale for the STKs turns out to be the best approach. Automatic scale selection performs closely and can be a good option if the subject's height cannot be measured due to occlusions. Constant scale for all subjects performs the worst. However, the performance of our algorithm is better than existing techniques in all three settings. 

\begin{table} 
\caption{\small Average recognition accuracy of the proposed method in three different settings on the Northwestern-UCLA Multiview Action3D\cite{and-or} and the UWA3D Multiview Activity II datasets. (1) Constant spatial scale for all subjects, (2) ratio of subject's height as the spatial scale, and (3) automatic spatial scale selection\cite{Mian}.}
    \centering \begin{tabular}{lcccccc}
    \toprule
    \multirow{2}{*}{Dataset} & \multicolumn{3}{c}{Spatial scale selection method} \\
    \cline{2-4} & {Constant} & {Subject height} & {Automatic} \\
    \midrule
    \midrule
    N-UCLA  & 77.9 & {\bf 80.0} & 79.5\\
    UWA3DII & 48.0 & {\bf 52.2} & 50.9 \\
    \bottomrule
    \end{tabular}
  \label{tab:spatialscale}
\end{table}

\subsubsection{Automatic Temporal Scale Selection}
We evaluate the improvement gained by our method using automatic temporal scale selection by repeating our experiments with constant temporal scale for STK detection and Local HOPC descriptor extraction.  Table~\ref{tab:Speed} shows the average recognition accuracy of our proposed method using a constant temporal scale ($\tau=2$) and automatic temporal scale selection. The proposed automatic temporal scale selection technique achieved higher accuracy which shows the robustness of our method to action speed variations. 

\begin{table} 
\caption{\small Average recognition accuracy of the proposed method in two different settings on  the Northwestern-UCLA Multiview Action3D\cite{and-or} and the UWA3D Multiview Activity II datasets. (1) Constant temporal scale ($\tau=2$) and (2) our automatic temporal scale selection technique.}
    \centering \begin{tabular}{lcccccc}
    \toprule
    \multirow{2}{*}{Dataset} & \multicolumn{2}{c}{Temporal scale selection method} \\
    \cline{2-3} & {Constant} & {Automatic} \\
    \midrule
    \midrule
    N-UCLA  & 78.3 & {\bf 80.0} \\
    UWA3DII & 49.2 & {\bf 52.2} \\
    \bottomrule
    \end{tabular}
  \label{tab:Speed}
\end{table}

\subsection{Evaluation of Parameters and Computation Time}
\label{Parameters}
\subsubsection{Number of STKs}
\begin{figure}[t]
\begin{center}
\includegraphics[width=\linewidth]{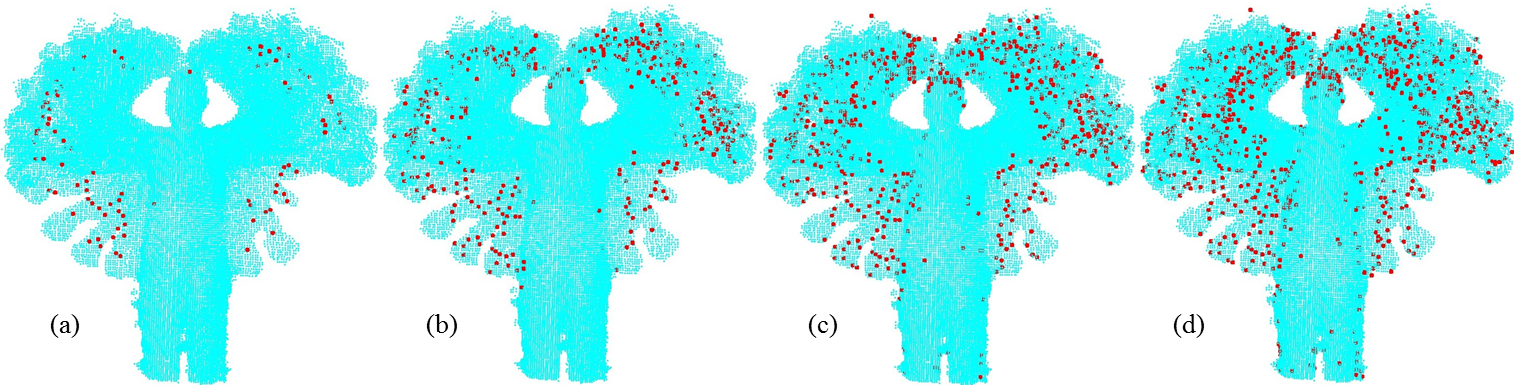}
\end{center}
\caption{\small STKs (shown in red) extracted using our proposed detector. STKs are projected on $XYZ$ dimensions of all points within a 3D pointcloud sequence corresponding to the action {\it two hand waving}. The top $n_k=100, 400, 700, 1000$ with the best quality are shown in (a)-(d), respectively. Note that the highest quality STKs are detected where significant movement is performed. Noisy points begin to appear as late as $n_k=1000$.}
\label{fig:NoSTK}
\end{figure}

\begin{figure}
\begin{center}
\includegraphics[width=\linewidth]{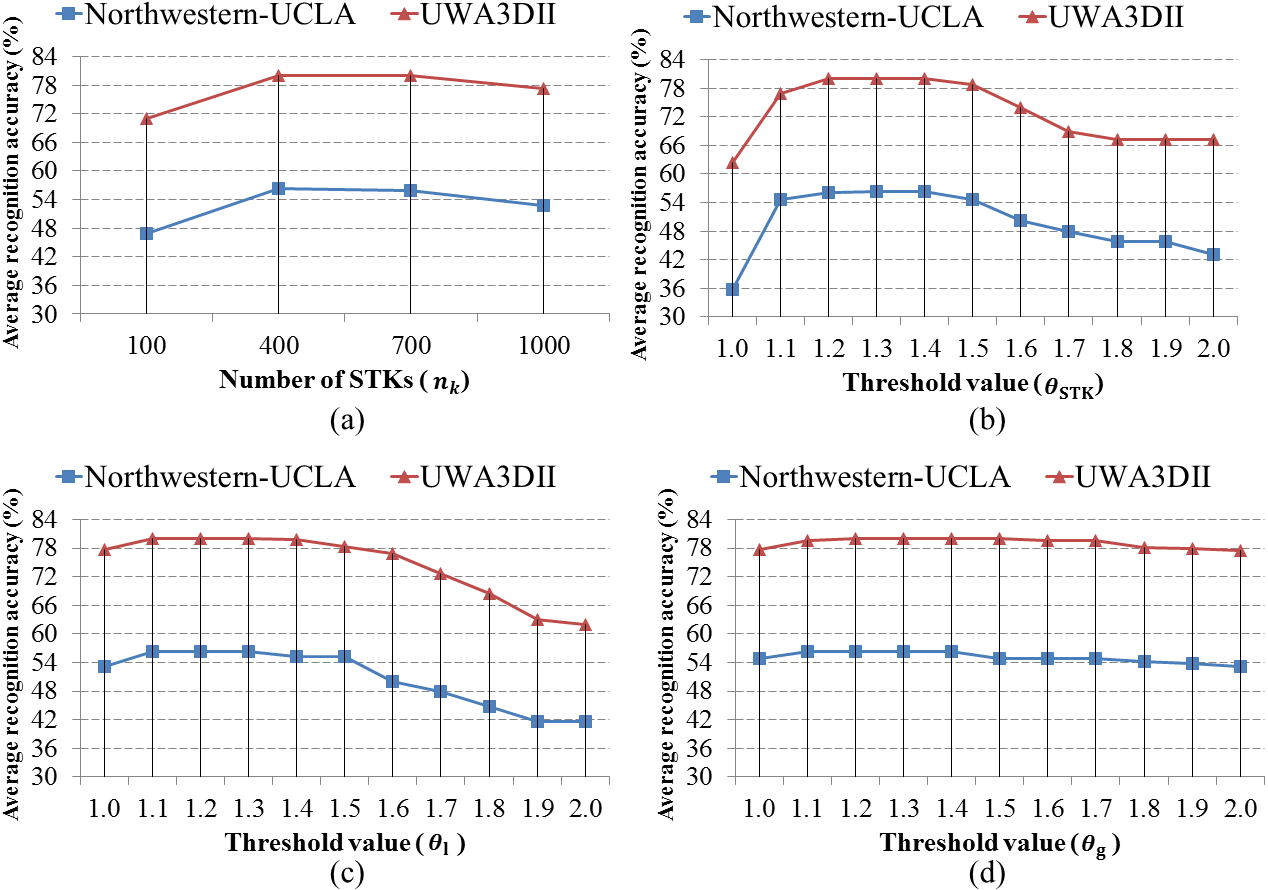}
\end{center}
\caption{\small Average recognition accuracy of Local \mbox{HOPC+STK-D} versus (a) the number of STKs, (b) $\theta_{\text{STK}}$, (c) $\theta_{\text{l}}$, and (d) $\theta_{\text{g}}$ on the Northwestern-UCLA\cite{and-or} and the UWA3D Multiview Activity II datasets.}
\label{fig:plot}
\end{figure}

To study the effect of the total number of STKs ($n_k$), we select STKs with the top $n_k=100, 400, 700, 1000$ {\it quality factors} as shown in  Fig.~\ref{fig:NoSTK}. Note how the STK detector effectively captures the movement of the hands in the highest quality STKs, and noisy points begin to appear as late as $n_k=1000$. Figure~\ref{fig:plot}(a) shows the influence of the number of STKs on the average recognition accuracy. The proposed method achieves the best recognition accuracy when $n_k=400$; however, the performance remains stable up to $n_k=700$.

\subsubsection{Threshold Values}
We evaluate the effect of the eigenratio thresholds $\theta_{\text{STK}}$ for STK detection in~\eqref{eq:eigenratio}, $\theta_{\text{l}}$ for view-invariant Local HOPC in Section~\ref{ModuleC}, and $\theta_{\text{g}}$ for STK-D in Section~\ref{ModuleD} on the average recognition accuracy of our proposed method. Figures~\ref{fig:plot}(b)-(d) show our results. Notice that there is a large range ($1.1 \le \theta_{\text{STK}} \le 1.5$) over which the recognition accuracy remains stable. For very small values of $\theta_{\text{STK}}$, a unique coordinate basis can not be obtained and for larger values of $\theta_{\text{STK}}$, the number of detected STKs is not sufficient.

A more stable trend in recognition accuracy can be observed for varying the thresholds $\theta_{\text{l}}$ and $\theta_{\text{g}}$. The recognition accuracy starts to decrease when $\theta_{\text{l}} > 1.5$ because the number of points within the spatio-temporal support volume of STKs which have unique eigenvectors starts to decrease. Finally, varying $\theta_{\text{g}}$ does not change the accuracy significantly because the extracted STKs from most actions already have a unique orientation and the proposed iterative refinement process (Section~\ref{ModuleD}) almost always finds an unambiguous orientation for all values of $\theta_{\text{g}}$.



\subsubsection{Computation Time}
The average computational time of the STK detection is $1.7$ seconds per frame on a $3.4$GHz machine with $24$GB RAM using Matlab. The calculation of Local HOPC at STKs takes $0.2$ seconds per frame. The overall computational time of the proposed view-invariant method is about $2$ seconds per frame. However, the proposed view-dependent Holistic HOPC is faster and take only $0.6$ seconds per frame. The average computation time of the nearest competitor AOG\cite{and-or} that uses depth images is $1.4$ seconds per frame. However, our method outperforms AOG\cite{and-or} on single-view and multi-view datasets by significant margins. Moreover, the calculation of STK and HOPC are individually parallel in nature and can be implemented on a GPU.

\section{Discussion and Conclusion}
Performance of the current 3D action recognition techniques degrades under viewpoint variations because they treat 3D videos as depth image sequences. Depth images are defined with respect to a particular viewpoint and are thus highly dependent on the viewpoint. We have proposed an algorithm for cross-view action recognition which directly processes 3D pointcloud videos to achieve robustness to variations in viewpoint, subject scale and action speed. We have also proposed the HOPC descriptor that is well integrated with our proposed spatio-temporal keypoint (STK) detection algorithm. Local HOPC descriptor combined with global STK-Distribution achieve state-of-the-art results on two standard cross-view action recognition datasets.

Unlike HOJ3D\cite{ViewInvariantJoint3D}, LARP\cite{LARP}, AE\cite{ActionLet2012}, and AOG\cite{and-or}, our method does not require skeleton data. Skeleton or joint estimation methods such as\cite{SingleDepth} do not work well when the human body is only partially visible. Moreover, joint estimation may not be reliable when the subject is not in an upright position (e.g. patient lying on bed)\cite{DSTIP} or touches the background. Finally, surveillance cameras are usually mounted at elevated angles which causes further difficulties in joint estimation\cite{DSTIP}. Thus, our proposed methods (and other non-skeleton based methods) are more general in the sense that they can be applied to a wider variety of action recognition problems where skeletonization of the data is either not possible or has not been achieved yet.

\section*{Acknowledgment}
\addcontentsline{toc}{section}{Acknowledgment}
We thank the authors of\cite{and-or} for providing the Northwestern-UCLA Multiview Action3D dataset  and especially Dr Jiang Wang for answering our questions about the implementation of AOG\cite{and-or} and AE\cite{AL2} methods. We also thank the authors of\cite{CCD,LARP,HON4D,SNV,DSTIP, virtualviews,NDVV} for making their codes publicly available. This research is supported by ARC Discovery grant DP110102399.

\ifCLASSOPTIONcaptionsoff
  \newpage
\fi



%
\bibliographystyle{IEEEtran}
\bibliography{BibECCV8a}

%


\begin{IEEEbiography}[{\includegraphics[width=1in,height=1.25in,clip,keepaspectratio]{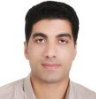}}]
{Hossein Rahmani} received his B.Sc. degree in Computer Software Engineering in 2004 from Isfahan University of Technology (IUT), Isfahan, Iran and the M.S. degree in Software Engineering in 2010 from Shahid Beheshti University (SBU), Tehran, Iran. His research interests include computer vision, 3D shape analysis, and pattern recognition. He is currently working towards his PhD degree in computer science from The University of Western Australia. His current research is focused on RGB-Depth based human activity recognition.
\end{IEEEbiography}


\begin{IEEEbiography}[{\includegraphics[width=1in,height=1.25in,clip,keepaspectratio]{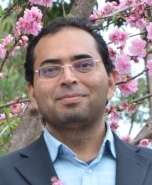}}]
{Arif Mahmood} obtained Gold Medal in MS and completed PhD with distinction from Lahore University of Management Sciences in 2011. He is currently PostDoc Researcher in Qatar University. Previously he was Research Assistant Professor in The University of Western Australia. His major research interests include data clustering and classification, action and object recognition, crowd analysis, community detection and content based image search and matching. 
\end{IEEEbiography}


\begin{IEEEbiography}[{\includegraphics[width=1in,height=1.25in,clip,keepaspectratio]{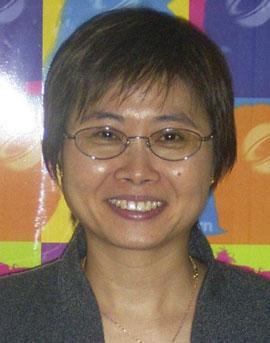}}]
{Du Huynh} is an Associate Professor at the School of Computer Science and Software Engineering, The University of Western Australia. She obtained her Ph.D in Computer Vision, in 1994, at the same university. Since then, she has worked for the Australian Cooperative Research Centre for Sensor Signal and Information Processing (CSSIP) and Murdoch University. She has been a visiting scholar at Lund University, Malmo University, Chinese University of Hong Kong, Nagoya University, Gunma University, and the University of Melbourne. Associate Professor Huynh has won several grants funded by the Australian Research Council. Her research interests include shape from motion,
multiple view geometry, video image processing, and visual tracking.
\end{IEEEbiography}


\begin{IEEEbiography}[{\includegraphics[width=1in,height=1.25in,clip,keepaspectratio]{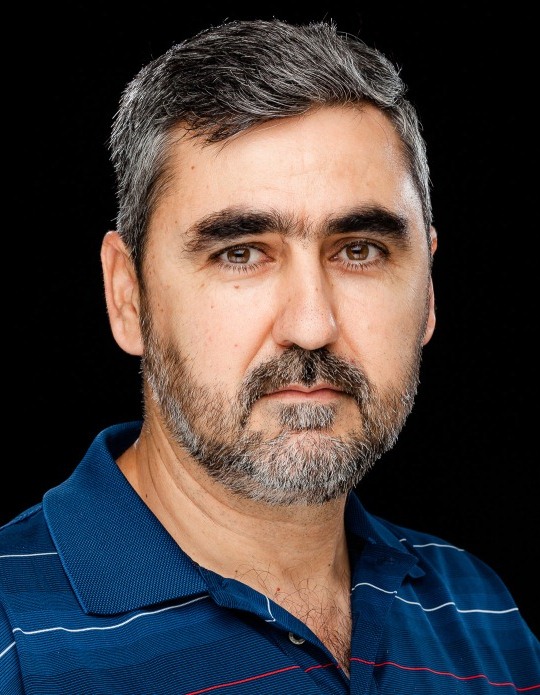}}]
{Ajmal Mian} completed his PhD from The University of Western Australia in 2006 with distinction and received the Australasian Distinguished Doctoral Dissertation Award from Computing Research and Education Association of Australasia. He received two prestigious nationally competitive fellowships namely the Australian Postdoctoral Fellowship in 2008 and the Australian Research Fellowship in 2011. He received the UWA Outstanding Young Investigator Award in 2011, the West Australian Early Career Scientist of the Year award in 2012 and the Vice-Chancellor’s Mid-Career Research Award in 2014. He has secured five Australian Research Council grants worth over \$2.3 Million. He is currently with the School of Computer Science and Software Engineering, The University of Western Australia. His research interests include computer vision, action recognition, 3D shape analysis, hyperspectral image analysis,  machine learning, and multimodal biometrics.

\end{IEEEbiography}

%
%
%
%



\end{document}